\documentclass[lettersize,journal]{IEEEtran}
\usepackage{amsmath,amsfonts}
\usepackage{algorithmic}
\usepackage{array}
\usepackage{stfloats}
\usepackage{url}
\usepackage{verbatim}
\usepackage{graphicx}
\usepackage{float}

\usepackage[T1]{fontenc}
\usepackage{booktabs}
\usepackage{multirow} 
\usepackage{subfigure}
\usepackage{verbatim}
\usepackage{cite}
\usepackage{afterpage}
\usepackage{pifont}
\usepackage[ruled,linesnumbered]{algorithm2e}
\usepackage[colorlinks,
linkcolor=blue,
anchorcolor=blue,
citecolor=blue]{hyperref}

\usepackage{threeparttable}

\def\BibTeX{{\rm B\kern-.05em{\sc i\kern-.025em b}\kern-.08em
    T\kern-.1667em\lower.7ex\hbox{E}\kern-.125emX}}
\usepackage{balance}

\title{PG-SLAM: Photo-realistic and Geometry-aware RGB-D SLAM in Dynamic Environments}
\author{Haoang Li, Xiangqi Meng, Xingxing Zuo, Zhe Liu, Hesheng Wang, and Daniel Cremers
\thanks{H. Li and X. Meng are with the Thrust of Robotics and Autonomous Systems, The Hong Kong University of Science and Technology (Guangzhou), Guangzhou, China.}
\thanks{
X. Zuo and D. Cremers are with the School of Computation, Information and Technology, Technical University of Munich, Munich, Germany.
}
\thanks{
	Z. Liu is with the MoE Key Lab of Artificial Intelligence, Shanghai Jiao
	Tong University, Shanghai, China.
}
\thanks{
	H. Wang is with the Department of Automation, Key Laboratory of System Control and Information Processing of Ministry of Education, Shanghai Jiao Tong University, Shanghai, China.
}
}

\begin{document}

\maketitle

\begin{abstract}

Simultaneous localization and mapping (SLAM) has 
achieved impressive performance in static environments. However, SLAM in dynamic environments remains an open question. Many methods directly filter out dynamic objects, resulting in incomplete scene reconstruction and limited accuracy of camera localization.
The other works express dynamic objects by point clouds, sparse joints, or coarse meshes, which fails to provide a photo-realistic representation.
To overcome the above limitations, we propose a photo-realistic and geometry-aware RGB-D SLAM method by extending Gaussian splatting. Our method is composed of three main modules to 1) map the dynamic foreground including non-rigid humans and rigid items, 2)
reconstruct the static background, and 3) localize the camera.
To map the foreground, we focus on modeling the deformations and/or motions. We consider the shape priors of humans and exploit geometric and appearance constraints
of humans and items. For background mapping, we design an optimization strategy between neighboring local maps by integrating appearance constraint into geometric alignment. As to camera localization, we leverage both static background and dynamic foreground to increase the observations for noise compensation.
We explore the geometric and appearance constraints by associating 3D
Gaussians with 2D optical flows and pixel patches. Experiments on various real-world datasets demonstrate that our method outperforms state-of-the-art approaches in terms of camera localization and scene representation. Source codes will be publicly available upon paper acceptance.
\end{abstract}


\section{Introduction}
Simultaneous localization and mapping (SLAM) is a crucial technique to help mobile agents autonomously navigate in unknown environments. It has various applications such as robotics, autonomous vehicles, and augmented reality~\cite{cadena2016past}.
Among different types of SLAM configurations, visual SLAM based on RGB and/or depth cameras gain popularity thanks to low cost and high compactness. Previous visual SLAM research develops various solutions based on the extended Kalman filter~\cite{MonoSlam}, feature correspondences~\cite{mur2017orb}, and direct pixel alignment~\cite{dso}. 
They achieve reliable camera localization, but fail to provide photo-realistic scene representation based on point clouds. To improve scene representation,
some new approaches based on the differentiable rendering~\cite{sucar2021imap,zhu2022nice,yan2024gs} were proposed. They feature high-quality scene rendering from novel views and gained wide attention in recent years.

The above methods mainly focus on static environments, regardless of the strategies for scene representation. In practice, dynamic environments are very common, exemplified by crowded shopping malls and busy roads.
A dynamic environment is typically composed of static background such as walls and floors, and dynamic foreground such as non-rigid humans and rigid items. SLAM in dynamic environments is challenging since dynamic foreground does not satisfy some basic geometric constraints such as epipolar geometry~\cite{Hartley2004}, and thus significantly affects the system robustness.
To overcome this challenge, 
a straightforward solution is 
filtering out dynamic foreground and only using static background for camera localization~\cite{sun2018motion,yu2018ds,jiang2024rodyn}.  This solution is not ideal  for two main reasons. First, it fails to map the foreground, resulting in an incomplete scene reconstruction.
Second, the accuracy of camera localization is affected by the loss of foreground information,
especially when dynamic foreground is dominant in an image.

Contrary to the above foreground filtering-based solution,
some works involve dynamic foreground in camera localization and scene mapping.
Early methods~\cite{zhang2020vdo,bescos2021dynaslam} only focus on rigid items like cars via exploring some geometric relationship.
Recent approaches~\cite{qiu2022airdos,henning2022bodyslam} further model the motions and shapes of non-rigid humans based on prior constraints.
However, the mapping results of the above methods 
lack fine details and textures. For example, the rigid item-oriented methods can only generate dynamic point clouds. The approaches targeted at non-rigid humans merely 
reconstruct
sparse joints or coarse meshes.
To overcome the above limitations, we propose a photo-realistic and geometry-aware
RGB-D SLAM method named PG-SLAM in dynamic environments.   As shown in Fig.~\ref{fig:overview},
our method is composed of three main modules
to 1) reconstruct the dynamic foreground including non-rigid humans and rigid items, 2)
map the static background, and 3) localize the moving camera.
All these modules are designed by extending Gaussian splatting (GS)~\cite{kerbl3Dgaussians}, a recently popular differentiable rendering technique.

To achieve a photo-realistic mapping of dynamic foreground, we propose dynamic GS constrained by geometric priors and human/item motions.
 For non-rigid humans, we adapt GS in three aspects. First, we attach Gaussians to the skinned multi-person linear (SMPL) model~\cite{SMPL}, which satisfies the articulated constraints of humans.
Second, we design a neural network to model the deformation of humans over time by considering the varied human pose.
Third, to compute the motions of humans, we exploit the 3D root joint of the SMPL model
and regularize the scales of Gaussians.
As for the mapping of rigid items, 
we focus on computing the rigid motions of these items.
To achieve this, we associate dynamic Gaussians at different times based on the optical flows, followed by aligning these Gaussinas. 
Moreover, we improve the completeness of 
mapping by effectively managing the previously and newly observed  parts of items.

As to the mapping of static background, given that an area can be consistently observed by multiple sequential images, 
we introduce a local map to manage these images.
Gaussian optimization within such a local map improves the accuracy thanks to multiple-view 
	 appearance constraints.
	 Moreover, we propose an optimization strategy between neighboring local maps
 based on both geometric and appearance constraints.
	The geometric constraint is formulated as an iterative alignment between 
 the centers of
 Gaussians from neighboring local maps.
	Moreover, we integrate appearance constraint into each iteration,
leading to more reliable convergence.
	The above  optimization strategy can reduce the accumulated error and ensure consistency between local maps.
	In addition, this strategy can be applied to loop closure when two local maps are not temporally continuous but spatially overlapping.

For camera  localization, our method leverages information of not only static background but also dynamic foreground. Additional observations can compensate for noise and thus improve the localization accuracy. To achieve this, we propose a two-stage strategy to estimate the camera pose in a coarse-to-fine manner based on both appearance and geometric constraints. 
For one thing, we use the camera pose to render Gaussians as images, formulating the appearance constraints.
For another, we use the camera pose to project Gaussians and blend them based on the rendering weights, generating the projected optical flows.
By aligning the observed and projected optical flows, we define the geometric constraint.
By combining the above 
constraints, our method achieves high-accuracy camera localization.

Overall, we propose a photo-realistic and geometry-aware RGB-D SLAM method for dynamic environments.
Our main contributions are summarized as follows:
\begin{itemize}
\item To the best of our knowledge, we propose the first Gaussian splatting-based SLAM method that can 
not only localize the camera and reconstruct the static background, but also map the dynamic humans and items. 
\item Our method can provide photo-realistic representation of dynamic scenes. For foreground mapping, we 
consider
shape priors of humans and exploit geometric and appearance constraints with respect to Gaussians.
To map the background, we design an effective optimization strategy between neighboring local maps.
	\item 
 Our method simultaneously uses the geometric and appearance constraints to localize the camera by associating 3D Gaussians with 2D optical flows and pixel patches.
We leverage information of both static background and dynamic foreground to compensate for noise, effectively improving the localization accuracy.
\end{itemize}
Experiments on various real-world datasets demonstrate that our method outperforms state-of-the-art approaches.

\section{Related Work}
\label{sec:rel_work}
Existing visual SLAM methods can be roughly classified into three categories in terms of representation strategies, i.e. the traditional representation-based, the implicit representation-based, and the GS-based approaches. 
We review these types of work for both static and dynamic environments.

\subsection{Traditional Representation-based Methods}

Point cloud is one of the dominant strategies for 3D representation. To reconstruct point clouds, previous SLAM methods match 2D point features for triangulation~\cite{mur2017orb,MonoSlam} or back-project depth images~\cite{whelan2016elasticfusion,RGBD-Tro}.
These methods also use point clouds to localize the camera based on several geometric algorithms.
 Mesh and voxel are the other common 3D representation strategies. Some SLAM approaches~\cite{slam_mesh,slam_voxel} consider these types of data to 
 achieve a continuous representation, but partly increase the complexity of map update.
 The above SLAM methods can reliably work in static environments,
 but become unstable
in the presence of dynamic objects due to lack of effective geometric constraints. 

To improve the SLAM robustness in dynamic scenes, a straightforward strategy is detecting and filtering out dynamic objects.
To achieve this, FlowFusion \cite{zhang2020flowfusion} 
utilizes the optical flows, DSLAM~\cite{Pami_Point} considers the temporal correlation of points, and DS-SLAM \cite{yu2018ds} leverages the semantic information.
While these methods can provide satisfactory estimation of camera trajectory, they are unable to reconstruct dynamic objects.
By contrast, some methods were designed to map dynamic objects.
For example, VDO-SLAM \cite{zhang2020vdo} and DynaSLAM II~\cite{bescos2021dynaslam} 
explore geometric constraints with respect to the motions 
of rigid items.
 AirDOS~\cite{qiu2022airdos} extends the rigid items to non-rigid humans
by expressing humans with a set of articulated joints.
By contrast,
Body-SLAM~\cite{henning2022bodyslam} can additionally provide coarse human shapes based on the SMPL model.
The main limitation of the above methods is that they neglect the details and textures of dynamic objects and thus fail to provide a photo-realistic scene representation.

\subsection{Implicit Representation-based Methods}
Signed distance function fields~\cite{Park_2019_CVPR} and  neural radiance fields (NeRF)~\cite{mildenhall2020nerf} are representative implicit scene representation methods. They express a 3D environment based on neural networks whose inputs are 3D coordinates of arbitrary positions and outputs are geometric and appearance attributes of these positions. They can provide differentiable and photo-realistic rendering of 3D scenes 
from novel views.
These techniques have already been used for SLAM to jointly represent the 3D scene and optimize the camera poses~\cite{sucar2021imap,zhu2022nice,yang2022vox,johari2023eslam,wang2023co,zhang2023go,sandstrom2023point}. 
The seminal work iMAP~\cite{sucar2021imap} adapts the original NeRF by neglecting 
view directions to 
efficiently express a 3D space.
A subsequent method NICE-SLAM~\cite{zhu2022nice} introduces the hierarchical voxel features as inputs to improve the generalization of scene representation.
The state-of-the-art work ESLAM \cite{johari2023eslam} leverages the multi-scale axis-aligned feature planes where the interpolated features are passed through a decoder to predict the truncated signed distance field. This method can generate more accurate 3D meshes than iMAP and NICE-SLAM.

The above methods are mainly designed for static scenes and 
can hardly handle dynamic environments.
To solve this problem, Rodyn-SLAM \cite{jiang2024rodyn} was recently proposed. This method
estimates the masks of dynamic objects and further eliminates these entities. 
While it achieves a relatively decent camera tracking performance, it cannot 
reconstruct dynamic objects. 
Moreover, 
the object filtering-based strategy
results in the loss of useful information, affecting the accuracy of camera localization.

\subsection{Gaussian Splatting-based Methods}
While the above NeRF has shown impressive performance in 3D scene representation, there is still room for improvement in quality and efficiency of rendering. GS
improves NeRF by replacing the implicit neural network with a set of explicit Gaussians. 
 Several concurrent works~\cite{GS-SLAM,splatam,yan2024gs,gaussianslam} integrated this technique into SLAM. They render Gaussians as RGB and depth images, and employ the photometric and depth losses to jointly optimize Gaussians and camera pose. They also tailor GS to special configurations of SLAM, such as sequential images with relatively similar views.
For example,
MonoGS~\cite{GS-SLAM} introduces the isotropic loss to avoid the ellipsoids with too long axes.
SplatSLAM \cite{splatam} proposes a silhouette-guided pixel selection strategy to exploit reliable pixels for optimization.
These methods typically achieve more accurate 3D reconstruction and/or faster camera localization than the above NeRF-based approaches. 

Despite high reliability in static environments,
the performance of
the above methods is unsatisfactory in dynamic scenes. 
The reason is that they use static Gaussians to express dynamic object movement,  resulting in blurry rendered images that drastically affect  
SLAM optimization.
To solve this problem,
a very recent work~\cite{DG-SLAM} follows the above Rodyn-SLAM to filter out dynamic objects. However, it leads to an incomplete scene reconstruction.

Overall, existing visual SLAM methods 
fail to achieve a photo-realistic and complete scene representation, as well as accurate camera localization in dynamic environments.
We overcome these limitations by proposing 
dynamic GS constrained by geometric priors and human/item motions. Our  optimization strategies based on both appearance and geometric constraints effectively improve the accuracy of SLAM.

\begin{figure*}[!t] 
    \centering
    \includegraphics[width=1.0\textwidth]{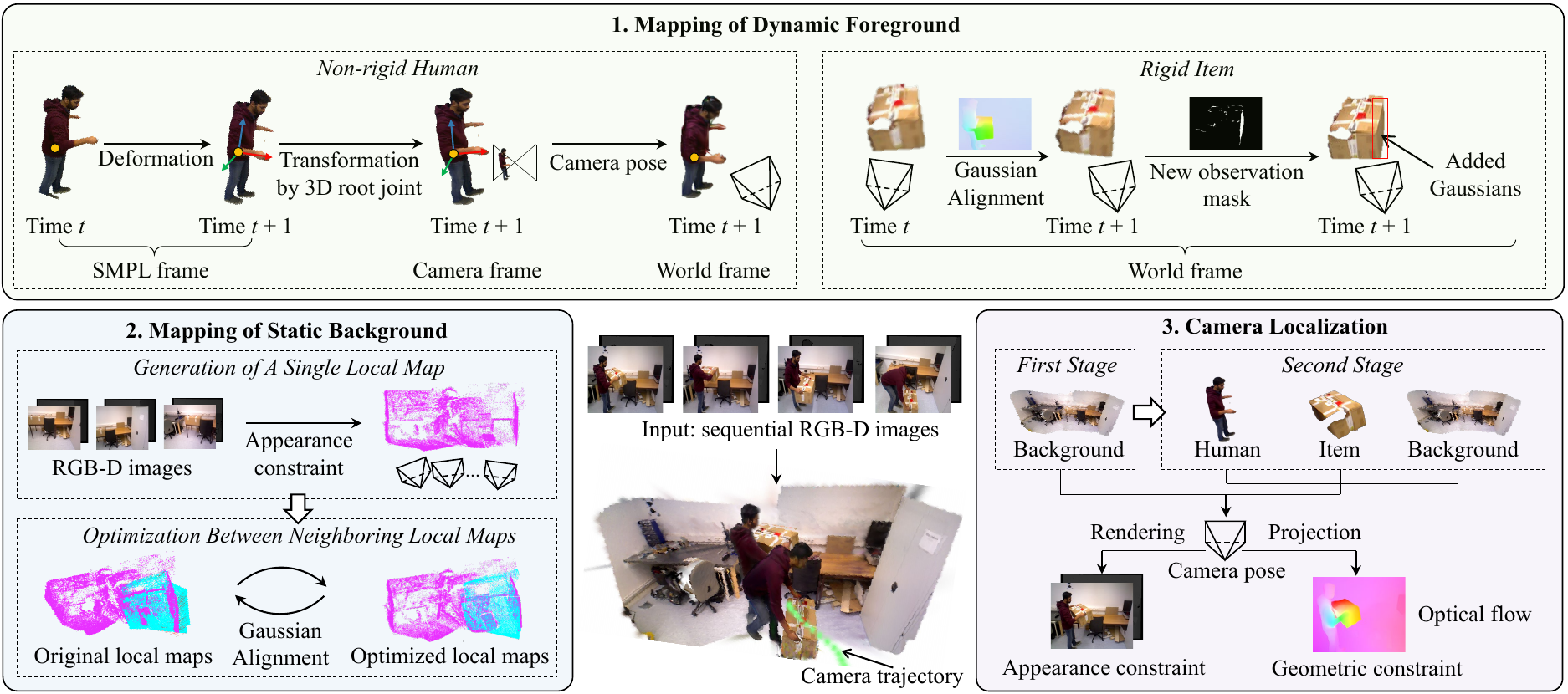} 
    \caption{Overview of our SLAM method. Given sequential RGB-D images obtained in a dynamic environment, our method can not only reconstruct the static background and localize the camera, but also map the dynamic foreground. By optimizing Gaussians based on appearance and geometric constraints, our method can provide photo-realistic scene representation and accurate camera localization.}
    \label{fig:overview}
\end{figure*}
\section{Problem Formulation}

In this section, we first introduce preliminary knowledge of GS and SMPL model, and then present an overview of our SLAM system. 

\subsection{Preliminary}

\subsubsection{Gaussian Splatting} \label{label:subsub_pre_GS}
GS is a differentiable rendering technique, featuring higher performance and geometric explainability. 
Briefly, a 3D space is expressed by a set of Gaussians~$\mathcal{G}=\{G_i\}$, each of which is
defined by:
\begin{equation}
G_i(\mathbf{P})=o_i \cdot \exp  \big\{-\tfrac12(\mathbf{P}-\boldsymbol{\mu}_i)^{\top}\boldsymbol{\Sigma}_i^{-1}(\mathbf{P}-\boldsymbol{\mu}_i)  \big\},
\end{equation}
where $\mathbf{P}$ is an arbitrary position in 3D, $\boldsymbol{\mu}_i$,
$o_i$,
and $\boldsymbol{\Sigma}_i$ denote the center, opacity, and covariance matrix of the $i$-th Gaussian, respectively.
 A covariance matrix
 can be decomposed into the scale and rotation.
Each Gaussian is also associated with spherical harmonics to encode different colors along various view directions.

As to the differentiable rendering, a pixel~$\mathbf{p}$ and the camera center define a 3D projection ray. Along this ray, $N$ 3D Gaussians are
projected onto 2D image from close to far. The $i$-th projected 2D Gaussian is associated with 
the color~$c_i$, depth~$d_i$, and covariance~$\boldsymbol{\Sigma}_i^{\textnormal{2D}}$ ($1\leq i \leq N$). These parameters are
obtained by spherical harmonics, $z$-coordinate of the center, and covariance of the corresponding 3D Gaussian, respectively.
The above~$N$ 2D Gaussians are blended to determine 
the color~$c_{\mathbf{p}}$ and depth~$d_{\mathbf{p}}$ of the pixel~$\mathbf{p}$:
\begin{equation}
    c_{\mathbf{p}}=\sum_{i=1}^N c_i\alpha_i\prod_{j=1}^{i-1}\left(1-\alpha_j\right),\
    d_{\mathbf{p}}=\sum_{i=1}^N d_i\alpha_i\prod_{j=1}^{i-1}\left(1-\alpha_j\right),
\end{equation}
where $\alpha_i$ 
represents the blending weight of the $i$-th 2D Gaussian which is computed based on the opacity~$o_i$ and covariance~$\boldsymbol{\Sigma}_i^{\textnormal{2D}}$.
Through the above pixel-wise computation, the rendered RGB image~$\tilde{I}$ and depth image~$\tilde{D}$ can be obtained.
For symbol simplification, in our context, we denote the above rendering process by
$
 \tilde{I}, \tilde{D}  = \pi[ \mathcal{G} ]$.
 
 To measure the appearance similarity between the rendered and observed RGB images, the photometric error is 
defined by the combination of $L_1$ and SSIM errors~\cite{ssim}:
\begin{equation}
    L_{\textnormal{P}} = \lambda \cdot \|I - \tilde{I}\|_1 +
    (1-\lambda ) \cdot \big(1-\textnormal{SSIM}(I,\tilde{I}) \big),
\end{equation}
where $\| \cdot \|_1$ denotes $L_1$ error, 
$\lambda$ is the weight to control the trade-off between two terms.
In addition, to measure the similarity between the rendered and  observed depth images, we use the depth error~$L_{\textnormal{D}}$ defined by $L_1$ error.
The combination of photometric and depth errors forms the appearance constraint.
 
\subsubsection{SMPL Model} \label{subsubsec:smpl}

SMPL model~\cite{SMPL} can effectively express dynamic 3D humans. It is defined by a kinematic  skeleton composed of 24 body joints, the pose parameter~$\Theta$ to drive the joints, and the shape parameter~$\Phi$ to describe different heights and weights.
Body joints include a root joint and other ordinary joints.
The root joint is considered to be the pelvis, as this is the central part of the body from which the rest of the body's kinematic chain (spine, legs, arms, etc.) originates. 
The origin of the SMPL frame is located at the root joint.
The root joint is associated with the 
rotation and translation 
between the SMPL frame and the camera frame.
The $k$-th ordinary joint is associated with a transformation~$\theta^k$ that defines its position and orientation relative to its parent joint in the hierarchy.
These transformations~$\{ \theta^k \}$ constitute the human pose~$\Theta$. 
The human shape~$\Phi$ is defined as a low-dimensional vector.

\subsection{System Overview}
We begin with introducing basic setups. 
We follow existing SLAM methods~\cite{mur2017orb,sucar2021imap} to 
treat the first camera frame as the world frame.
The absolute camera pose~$\mathbf{T}_t$ represents the rotation and translation from the $t$-th camera frame to the world frame.
Given sequential RGB images, we use Mask R-CNN~\cite{maskrcnn} to segment non-rigid humans, rigid items, and static background.
Based on segmentation results,
we introduce several sets of Gaussians to express different parts of a dynamic environment.
Given a camera pose, each set of Gaussians is rendered as an RGB patch and a depth patch, instead of complete RGB-D images.
Without loss of generality, we consider the scene with a single human or item for illustration. Our method can be easily extended to environments with multiple dynamic objects, as will be shown in the experiments.

As shown in Fig.~\ref{fig:overview}, our SLAM method is composed of three main modules to 1) map the dynamic foreground, 2) map the static background, and 3) localize the camera.
Following dominant SLAM methods~\cite{splatam,
GS-SLAM}, our mapping modules rely on the rough camera pose estimated by the localization module. The localization module exploits Gaussians reconstructed by the mapping modules at the previous times.
\subsubsection{Mapping of Dynamic Foreground} 
For non-rigid humans, we attach Gaussians to the SMPL model and design a neural network to express the human deformation over time.
Then for human motion expression, we exploit the 3D root joint of the SMPL model to transform humans to the camera frame, and further use the camera pose to transform humans to the world frame. 
For rigid items, 
we  associate
dynamic Gaussians at different times based on the optical flows, followed by aligning these Gaussians to compute the item motions. Moreover, we add new Gaussians
based on a new observation mask.
Details are available in Section~\ref{sec:mapping_dyn_foreground}.

\subsubsection{Mapping of Static Background}
We generate a local map to manage multiple images that partially observe the same area. We optimize Gaussians within such a local map based on multiple-view appearance constraints. Moreover, we
propose an optimization strategy between neighboring local maps based on both geometric and appearance constraints. The
geometric constraint is formulated as an iterative alignment
between the centers of Gaussians from neighboring local maps. Moreover, we integrate appearance constraint into each
iteration.
Details are available in Section~\ref{sec:mapping_background}.

\subsubsection{Camera Localization}
We propose a two-stage strategy to estimate the camera pose in a coarse-to-fine manner. In the first stage, we only use the
static background. In the second stage, we exploit both static
background and dynamic foreground.
Our localization simultaneously leverages the appearance and geometric constraints with respect to the camera pose. 
For one thing, we use the camera pose to render Gaussians as
RGB-D patches, formulating the appearance constraints.
For another, we use the camera pose to project Gaussians, generating the projected optical flows. By aligning the observed and projected optical
flows, we define the geometric constraint.
Details are available in Section~\ref{sec:cam_localization}.

\section{Mapping of Dynamic Foreground}
\label{sec:mapping_dyn_foreground}
In this section, we present how we map dynamic foreground in the world frame. 
We aim to 1) model their 3D structures and 2) describe their motions and/or deformations.  We first introduce non-rigid humans, followed by rigid items.
\begin{figure}[!t]
	\centering
	\includegraphics[height=0.33\textwidth]{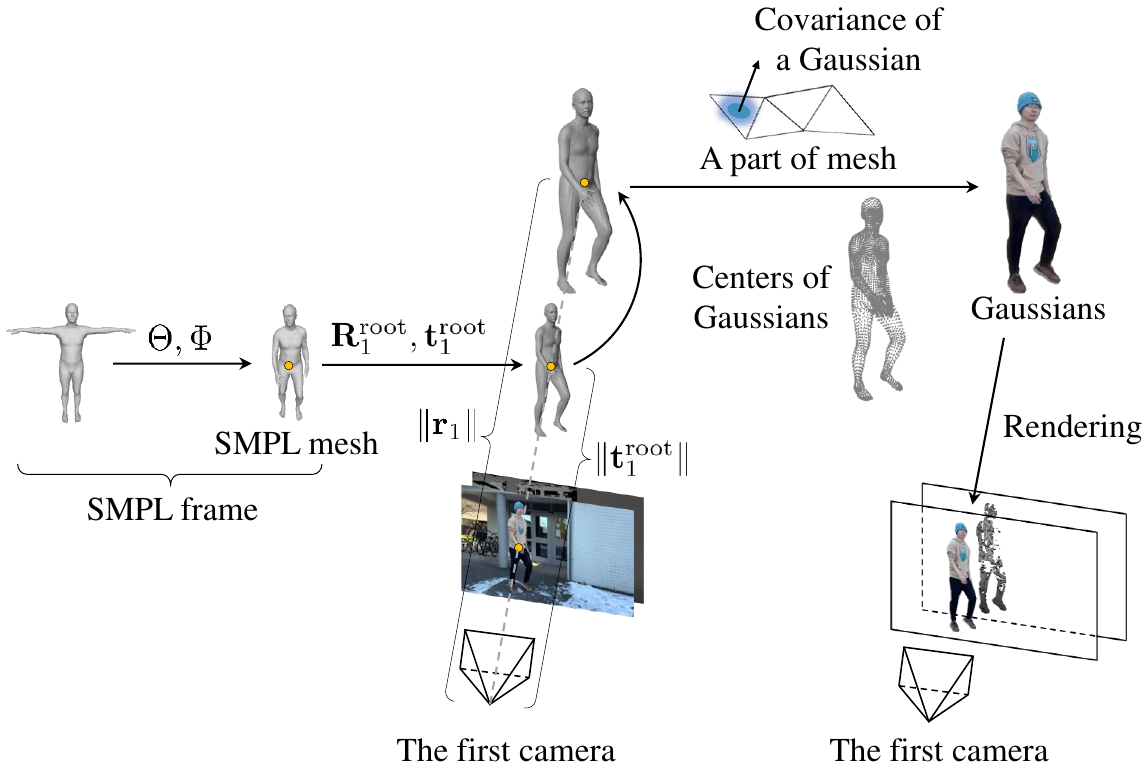}
	\caption{ Initialization of human Gaussians.
 Given the first RGB-D images, we generate an SMPL mesh up to scale, followed by transforming it to the camera frame based on the estimated transformation of root joint. We further determine the real scale of SMPL mesh based on the depth of root joint.
 Then we attach a set of Gaussians to the re-scaled SMPL mesh, and optimize these Gaussians based on the appearance constraint.
}
	\label{human_initialize}
\end{figure}

\subsection{Non-rigid Humans} 
\label{subsec:mapping_human}
Our 3D human representation is based on dynamic GS constrained by the SMPL model.
%
We first introduce the initialization of Gaussians, given the first RGB-D images. Then we present how we model the deformation and motion of Gaussians based on two neighboring RGB-D images.

\subsubsection{Initialization of Gaussians} \label{subsubsec:init}
As shown in Fig. \ref{human_initialize}, given the first RGB image, we employ ReFit~\cite{refit} to estimate the pose and shape parameters~$\Theta$ and $\Phi$. 
We use these parameters to
generate an SMPL mesh in the SMPL frame.
Based on the rotation~$\mathbf{R}^{\textnormal{root}}_1$  and translation~$\mathbf{t}^{\textnormal{root}}_1$  associated with the root joint,
we  
transform the generated SMPL mesh to the camera frame.
Note that the translation~$\mathbf{t}^{\textnormal{root}}_1$ is up-to-scale, i.e., the scale of SMPL mesh may not fit the real structure 
due to the inherent scale ambiguity of the perspective projection~\cite{refit}. 
To determine the real scale of the SMPL mesh,
we leverage the root joint.
Specifically,
ReFit provides the coordinates of the 2D root joint
in the first RGB image. 
We back-project this joint into 3D using its associated depth provided by the depth image, obtaining the 3D root joint~$\mathbf{r}_1$ in the camera frame. Recall that 3D root joint is the origin of the SMPL frame. Therefore,~$\mathbf{r}_1$  approximates to the real-scale translation from SMPL frame to the camera frame.
We compute the relative ratio between $\mathbf{r}_1$ and the above up-to-scale translation~$\mathbf{t}^{\textnormal{root}}_1$ 
by
$\| \mathbf{r}_1 \| / \| \mathbf{t}^{\textnormal{root}}_1 \|$, and use this ratio to re-scale the SMPL mesh.

Given the re-scaled SMPL mesh,
we attach a set of Gaussians  to its triangular facets. 
The center, scale, and rotation of a Gaussian can be roughly determined by the center, area, and normal of the corresponding facet, respectively.
Since SMPL mesh is typically not associated with color information, we randomize the opacity and spherical harmonics of Gaussians. 
To optimize these human Gaussians denoted by~$\mathcal{G}_1$,
we render them as 
RGB and depth patches:
$\tilde{I}_1(\mathcal{G}_1), \tilde{D}_1(\mathcal{G}_1) =
    \pi [\mathcal{G}_1]$. Then we use these rendered patches and 
    their corresponding observed patches
    to define the 
    photometric and depth 
    losses, optimizing the Gaussian~$\mathcal{G}_1$:
\begin{equation}
\min_{\mathcal{G}_{1} }   
\lambda_{\textnormal{P}} \cdot
 L_{\textnormal{P}}
\Big( \tilde{I}_1(\mathcal{G}_{1}) \Big)    + 
 \lambda_{\textnormal{D}} \cdot  L_{\textnormal{D}}  \Big(  \tilde{D}_1(\mathcal{G}_{1})  \Big),
\end{equation}
where  $\lambda_{\textnormal{P}}$ and $\lambda_{\textnormal{D}}$ denote coefficients of the photometric and depth losses, respectively.

\begin{figure}[!t]
	\centering
	\includegraphics[height=0.13\textwidth]{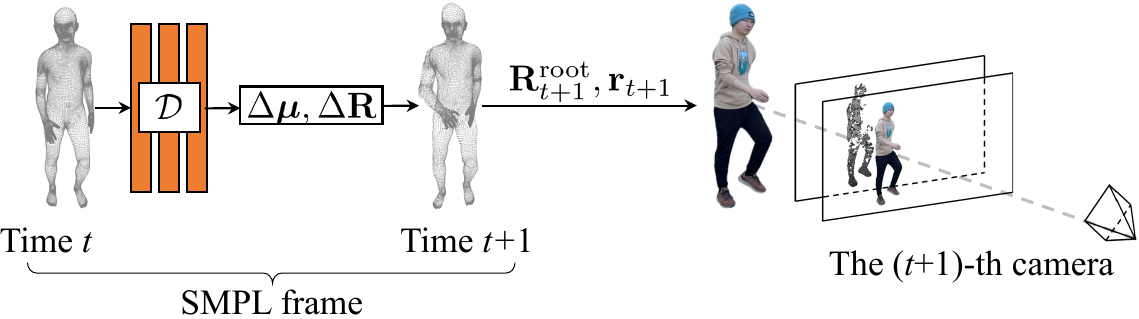}
 \caption{Update of human Gaussians.
 In the SMPL frame, we  use a neural network~$\mathcal{D}$ to deform Gaussians at time~$t$ into new Gaussians at time~$t+1$. Then we transform the deformed Gaussians to the $(t+1)$-th camera frame using the 
 transformation associated with
 the root joint. Finally, we optimize these Gaussians and the network~$\mathcal{D}$ based on the appearance constraint.}
	\label{fig:human_deform}
\end{figure}
\subsubsection{Update of Gaussians}
\label{subsubsec:human_update}
For a dynamic human, its associated Gaussians initialized above will be continuously updated over time.
Human typically exhibits two types of dynamics, i.e., deformation and transformation between times $t$ and $t+1$ ($t\geq1$). 
First, we estimate the deformation of Gaussians in the SMPL frame. 
We begin with using ReFit to estimate the human poses~$\Theta_t$ and $\Theta_{t+1}$ based on the RGB images~$I_t$ and $I_{t+1}$, respectively.
Intuitively, the variation of human pose leads to the variation of geometric attributes of Gaussians. As shown in Fig.~\ref{fig:human_deform}, to model this variation, 
we propose a neural network~$\mathcal{D}$.  
The input of this network is the variation of the human pose $\Delta \Theta_{t,t+1}$  from time $t$ to time $t+1$. 
It encodes the transformation variation of each ordinary joint (except for the root joint), i.e.,
$\Delta \Theta_{t,t+1} = [\Delta \theta^1_{t,t+1},\cdots,\Delta \theta^i_{t,t+1},\cdots,\Delta \theta^{23}_{t,t+1}]$.
 The transformation variation $\Delta \theta^i_{t,t+1}$ of the $i$-th ordinary joint is computed by
\begin{equation}
	\Delta \theta^i_{t,t+1} =  \prod \limits_{j \in \Omega_i}\theta_{t+1}^j (\prod \limits_{j \in \Omega_i}\theta_t^j)^{-1},
\end{equation}  
where $\Omega_i$   represents the set of parent joints of the $i$-th joint,
$\theta_t^j$ and
$\theta_{t+1}^j$ represent the transformations of the $j$-th parent joint at times $t$ and $t+1$ respectively.
Given the variation of the human pose~$\Delta \Theta_{t,t+1}$, the network~$\mathcal{D}$ can predict the variation in position~$\Delta \boldsymbol{\mu}_{t,t+1}$ and rotation~$\Delta \mathbf{R}_{t,t+1}$ of a Gaussian located at the position~$\boldsymbol{\mu}_t$
in the SMPL frame:
\begin{equation}
 	\Delta \boldsymbol{\mu}_{t,t+1}, \Delta \mathbf{R}_{t,t+1} = \mathcal{D} \big( \mathcal{E}(\boldsymbol{\mu}_t ),\Delta \Theta_{t,t+1} \big),
\end{equation}
where $\mathcal{E}(\cdot)$ denotes positional encoding.
The architecture of the network~$\mathcal{D}$ is based on the multi-layer perceptron (MLP).
Through this network, we can obtain the geometric attributes of  Gaussians at time $t+1$ in the SMPL frame by
\begin{equation}
    \boldsymbol{\mu}_{t+1} = \boldsymbol{\mu}_t + \Delta \boldsymbol{\mu}_{t,t+1},\
    \mathbf{R}_{t+1} =  \Delta \mathbf{R}_{t,t+1} \mathbf{R}_t.
\end{equation}
$\boldsymbol{\mu}_{t+1}$ and $\mathbf{R}_{t+1}$ belong to the attributes of  Gaussians  $\mathcal{G}_{t+1}$ at time~$t+1$. Therefore, $\mathcal{G}_{t+1}$ can be treated as the function with respect to the weights of the above network~$\mathcal{D}$, which is denoted by~$\mathcal{G}_{t+1}(\mathcal{D})$.
The other Gaussian attributes (color, opacity, and scale) are not changed by the network, i.e., they are shared by the Gaussians~$\mathcal{G}_t$ and~$\mathcal{G}_{t+1}$ at times~$t$ and $t+1$ respectively.

Second, we model the transformation of Gaussians in the world frame. 
We begin with 
expressing the status of Gaussians with respect to each camera frame.
Let us take the $(k+1)$-th camera frame for example.
Similar to the above Gaussian initialization, 
we use the RGB image~$I_{t+1}$ and depth image~$D_{t+1}$ to compute the rotation~$\mathbf{R}_{t+1}^{\textnormal{root}}$ and real-scale translation~$\mathbf{r}_{t+1}$  associated with the 3D root joint.\footnote{We do not consider the up-to-scale translation~$\mathbf{t}_{t+1}^{\textnormal{root}}$ associated with the 3D root joint since we have determined the real scale of SMPL mesh after initialization. In addition, the rotation~$\mathbf{R}_{t+1}^{\textnormal{root}}$ and real-scale translation~$\mathbf{r}_{t+1}$ can be optionally treated as the parameters to optimize, which contributes to improving the accuracy of human localization. 
In our experiments, we optimize these parameters.}
We use these parameters to transform the above deformed 
Gaussians~$\mathcal{G}_{t+1}(\mathcal{D})$
from the SMPL frame to the $(t+1)$-th camera frame. 
Then we render the transformed Gaussisans as 
RGB and depth patches:
$\tilde{I}_{t+1} \big( \mathcal{G}_{t+1}(\mathcal{D})  \big), \tilde{D}_{t+1} \big( \mathcal{G}_{t+1}(\mathcal{D}) \big) = \pi[\mathcal{G}_{t+1}(\mathcal{D}) ]$.
We use these rendered patches and their corresponding observed RGB-D patches 
to define the photometric and depth losses, optimizing the attributes (color, opacity, and scale) of Gaussians~$\mathcal{G}_{t+1}$
and the network~$\mathcal{D}$:
\begin{equation}
\min_{\mathcal{G}_{t+1}, \mathcal{D}   } 
\lambda_{\textnormal{P}} \cdot
 L_{\textnormal{P}}
\Big( \tilde{I}_{t+1}\big(\mathcal{G}_{t+1}(\mathcal{D}) \big) \Big) + 
\lambda_{\textnormal{D}} \cdot  L_{\textnormal{D}}  \Big(  \tilde{D}_{t+1}\big(\mathcal{G}_{t+1}(\mathcal{D}) \big)  \Big).
\end{equation}

At the same time, since the above partial attributes of Gaussians~$\mathcal{G}_{t+1}$  
are shared with Gaussians~$\mathcal{G}_{t}$,
we additionally consider the appearance constraints at time $t$.
We fix the exclusive parameters (centers and rotations) of Gaussians~$\mathcal{G}_t$ which have been optimized at the time~$t$, and optimize the shared parameters by rendering~$\mathcal{G}_{t}$ in the $t$-th camera frame. 
Therefore, the shared attributes  
are simultaneously constrained by two views, which ensures the temporal continuity of the Gaussian attributes and improves the reliability of Gaussian optimization. 
After optimizing Gaussians~$\mathcal{G}_{k+1}$ in the $(k+1)$-th camera frame, we use the camera pose $\mathbf{T}_{t+1}$ 
to transform these Gaussians 
to the world frame.
By repeating the above procedures, we can express the transformation of Gaussians in the world frame.

In addition, to improve the robustness of Gaussian optimization, we introduce a simple but effective human scale regularization loss. Specifically, the depth 
image 
is inevitably affected by noise.
Accordingly, the optimized 3D humans at different times may 
render plausible RGB appearance
but 
exhibit inconsistent sizes.
To solve this problem, we compute the average scale~$\hat{s}$ of all the  Gaussians initialized by the first image, 
and treat it as the constant standard scale.
At any time~$t$, we compute the average scale~$\bar{s}_t$ and force it to be similar to~$\hat{s}$. Based on this constraint, the sizes of the optimized 3D humans at different times fall into the same range.

\subsection{Rigid  Items} 
\label{subsec:obj_mapping}

Our 3D item representation is based on 
dynamic GS constrained by rigid transformation.
We first introduce the initialization of Gaussians, given the first RGB-D images. Then we present how we model the rigid transformation of Gaussians based on two neighboring images. Finally, given new RGB-D images, we introduce the 
addition of Gaussians.
\begin{figure}[!t]
	\centering
	\includegraphics[height=0.28\textwidth]{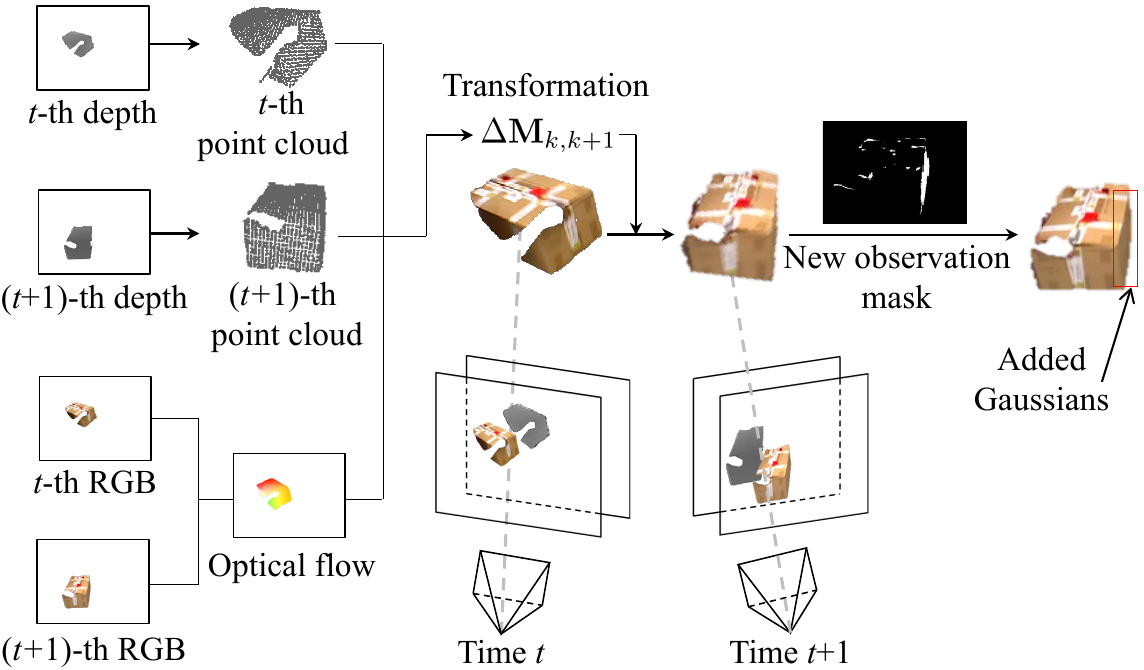}
	\caption{Rigid transformation and addition of item Gaussians. We first estimate the optical flow and back-project depth images to 
    establish 3D-3D point correspondences.
    Then we use these correspondences to roughly estimate the transformation,
    followed by optimizing Gaussians and transformation based on appearance constraint. Finally, we estimate the new observation mask that guides the addition of Gaussians using appearance constraint.}
	\label{fig:box}
\end{figure}

\subsubsection{Initialization of Gaussians}
\label{subsubsec:init_rigid}
Given the first RGB-D images, we back-project the pixels associated with the rigid item into 3D, obtaining a 3D point cloud in the camera frame. Then we use the position and color of each point to initialize the center and color of a Gaussian. The other attributes of these Gaussisans are randomized.
The optimization of these Gaussians is analogous to the above operations for humans. We render these Gaussians as RGB and depth patches, and define the photometric and depth losses for attribute optimization.

\subsubsection{Rigid Transformation of Gaussians}
Let us consider the $t$-th and $(t+1)$-th images to illustrate how we estimate the rigid transformation of Gaussians in a coarse-to-fine manner (see Fig.~\ref{fig:box}).
For the rough estimation, we first establish 2D-2D point correspondences in two RGB images by the optical flow estimation~\cite{raft}. Then we back-project the matched 2D pixels into 3D using their associated depths, obtaining two point clouds in respective camera frames. Based on the established 2D-2D point correspondences, 3D-3D point correspondences of two point clouds are known.
We further use the camera poses~$\mathbf{T}_t$ and~$\mathbf{T}_{t+1}$ to transform these two point clouds to the world frame, respectively. 
Finally, we compute the rigid transformation~$\Delta \mathbf{M}_{t,t+1}$ between two point clouds
using  
3D-3D correspondences via the singular value decomposition~\cite{icp_pami}.

To further improve the accuracy of the transformation~$\Delta \mathbf{M}_{t,t+1}$, we leverage the 
appearance constraint.
In the world frame, we first use $\Delta \mathbf{M}_{t,t+1}$ to transform the centers of Gaussians from time~$t$ to the time~$t+1$:
\begin{equation}
		\boldsymbol{{\mu}}^{\mathcal{W}}_{t+1} = \Delta \mathbf{M}_{t,t+1} (\boldsymbol{\mu}^{\mathcal{W}}_{t}).
\end{equation}
We further transform these centers to the $(k+1)$-th camera frame using the 
camera pose~$\mathbf{T}_{t+1}$.
Accordingly, Gaussians 
can be treated as a function with respect to the transformation $\Delta \mathbf{M}_{t,t+1}$, which is denoted by~$\mathcal{G}_{t+1}(\Delta \mathbf{M}_{t,t+1})$.
The other attributes of these Gaussians (color, opacity, rotation, and scale) are shared with Gaussians at time $t$.
We render Gaussians at time $t+1$
as RGB and depth patches:
$\tilde{I}_{t+1} \big(\mathcal{G}_{t+1}(\Delta \mathbf{M}_{t,t+1}) \big), \tilde{D}_{t+1} \big(\mathcal{G}_{t+1}(\Delta \mathbf{M}_{t,t+1}) \big) =
    \pi [\mathcal{G}_{t+1}(\Delta \mathbf{M}_{t,t+1})]$. Then we use these rendered patches and 
    their corresponding observed patches
    to define the 
    photometric and depth 
    losses, optimizing the attributes  of Gaussians~$\mathcal{G}_{t+1}$ (except for the center) and the transformation~$\Delta \mathbf{M}_{t,t+1}$:
\begin{equation}
\begin{split}
\min_{\mathcal{G}_{t+1}, \Delta \mathbf{M}_{t,t+1}  }  &
\lambda_{\textnormal{P}} \cdot
 L_{\textnormal{P}}
\Big( \tilde{I}_{t+1}\big(\mathcal{G}_{t+1}( \Delta \mathbf{M}_{t,t+1}  ) \big) \Big) + \\
& \lambda_{\textnormal{D}} \cdot  L_{\textnormal{D}}  \Big(  \tilde{D}_{t+1}\big(\mathcal{G}_{t+1}(  \Delta \mathbf{M}_{t,t+1} ) \big)  \Big).
\end{split}
\end{equation}

Moreover, similar to the above operation for humans, we additionally
consider the appearance constraints at time $t$.
We fix the centers of Gaussians~$\mathcal{G}_t$ which have been optimized at time~$t$, and optimize the shared parameters by rendering~$\mathcal{G}_{t}$ in the $t$-th camera frame. 
Therefore, the shared
attributes are simultaneously constrained by two views.

\subsubsection{Addition of Gaussians}
\label{subsub:add_gauss}

Due to viewpoint change and camera movement, some parts of 
a 3D item, which cannot be observed at time $t$, become visible at time $t+1$.
We aim to 
map these 3D parts, i.e., add some Gaussians to express these parts.
Given a new RGB image~$I_{t+1}$,
we first identify the pixels  that correspond to the newly observed but not reconstructed 3D parts of the item.
As shown in Fig.~\ref{fig:box}, 
we transform the above optimized
Gaussians~$\mathcal{G}_{t+1}$ to the $(t+1)$-th camera frame using the camera pose~$\mathbf{T}_{t+1}$, and render these Gaussians into a depth patch~$\tilde{D}_{t+1}$. 
If some pixels of the item in the RGB image $I_{t+1}$ are not associated with the rendered depths, we consider that these pixels
have not been associated with 3D Gaussians.
Accordingly, these pixels constitute the ``new observation mask''.

Then we back-project the pixels belonging to the mask into 3D based on their associated depths provided by the depth image~$D_{t+1}$, obtaining a set of colored 3D points in the $(t+1)$-th camera frame.
Then we use the positions and colors of these 3D points to initialize the centers and colors of
a set of new Gaussians.
 The optimization of these Gaussians is analogous with the above operation for initialization of item Gaussians. We
render these Gaussians as RGB and depth patches, and define
the photometric and depth losses for attribute optimization.
\section{Mapping of Static Background} \label{sec:mapping_background}
In this section, we present how we map static background.
To guarantee high accuracy,
we introduce local maps to reconstruct and manage Gaussians. 
Our main technical novelty lies in
an optimization strategy between neighboring local maps.
Different from related work~\cite{zhu2024loopsplat}, our method simultaneously leverages the geometric and appearance constraints, 
which can effectively reduce the accumulated error. 

\subsection{Generation of A Single Local Map}
Recall that to optimize the dynamic foreground at time~$t+1$, we only
use images obtained at times~$t+1$ and $t$. The reason is that the states of dynamic objects change quickly, and the previous images are not well-aligned to the current states.
By contrast, 
the attributes of background Gaussians remain consistent within a wider time window.
Therefore, we involve a larger number of sequential images in mapping a part of static background.
A set of Gaussians optimized by these images is called a local map.
In the following, let us consider a local map starting from time~$t$ for illustration.

\subsubsection{Initialization}
We initialize Gaussians following
the operations for rigid items (see Section~\ref{subsubsec:init_rigid}).
Briefly, given the RGB-D images~$(I_t, D_t)$ that is the first element of a local map, 
we generate a colored 3D point cloud and use it to initialize the
centers and colors of 
Gaussians. Then we render these Gaussians as RGB and depth patches, and exploit the photometric and depth losses for optimization.
After that, we use the camera pose to transform the optimized Gaussians from the $t$-th camera frame to the world frame.

Given the new RGB-D images~$(I_{t+m},D_{t+m})$ ($m \geq 1$),
we identify whether these images belong to the current local map.  If their associated camera pose does not significantly differ from that of the images~$(I_t,D_t)$, we treat~$(I_{t+m},D_{t+m})$ as a new element of the current local map. Otherwise, we initialize a new local map.

\subsubsection{Update} \label{Submap}

Assume that $(I_{t+m}, D_{t+m})$ is a new element of a local map. From this view, some parts of 3D background become newly observable. We follow the operation for rigid items to determine the new observation mask 
and add Gaussians associated with this mask
(see Section~\ref{subsub:add_gauss}).
Then we transform these Gaussians to the world frame~$\mathcal{W}$ and further append them to the local map.
After generating a local map~$\mathcal{G}^{\mathcal{W}}$ using $M$ images, we optimize 
this map based on multiple-view constraints.
For each of $M$ cameras, 
we use the $m$-th camera pose~$\mathbf{T}_m$ to transform Gaussians~$\mathcal{G}^{\mathcal{W}}$ from the world frame to the $m$-th camera frame and then render these Gaussians as RGB and depth patches: $\tilde{I}_m(\mathcal{G}^{\mathcal{W}}), \tilde{D}_m(\mathcal{G}^{\mathcal{W}}) =
    \pi [\mathcal{G}^{\mathcal{W}}, \mathbf{T}_m]$ ($1\leq m \leq M$). 
We use these rendered patches and 
    their corresponding observed patches
    to define the 
    photometric and depth 
    losses, optimizing the Gaussian~$\mathcal{G}^{\mathcal{W}}$:
\begin{equation}
\min_{\mathcal{G}^{\mathcal{W}} }  \sum_{m=1}^M
\lambda_{\textnormal{P}} \cdot
 L_{\textnormal{P}}
\Big( \tilde{I}_m(\mathcal{G}^{\mathcal{W}}) \Big)    + \lambda_{\textnormal{D}} \cdot  L_{\textnormal{D}} \Big(  \tilde{D}_m(\mathcal{G}^{\mathcal{W}}) \Big).
\end{equation}
\subsection{Optimization Between Neighboring Local Maps}
\label{subsec:opt_local_maps}
\begin{figure}[!t]
	\centering
	\includegraphics[width=0.4\textwidth]{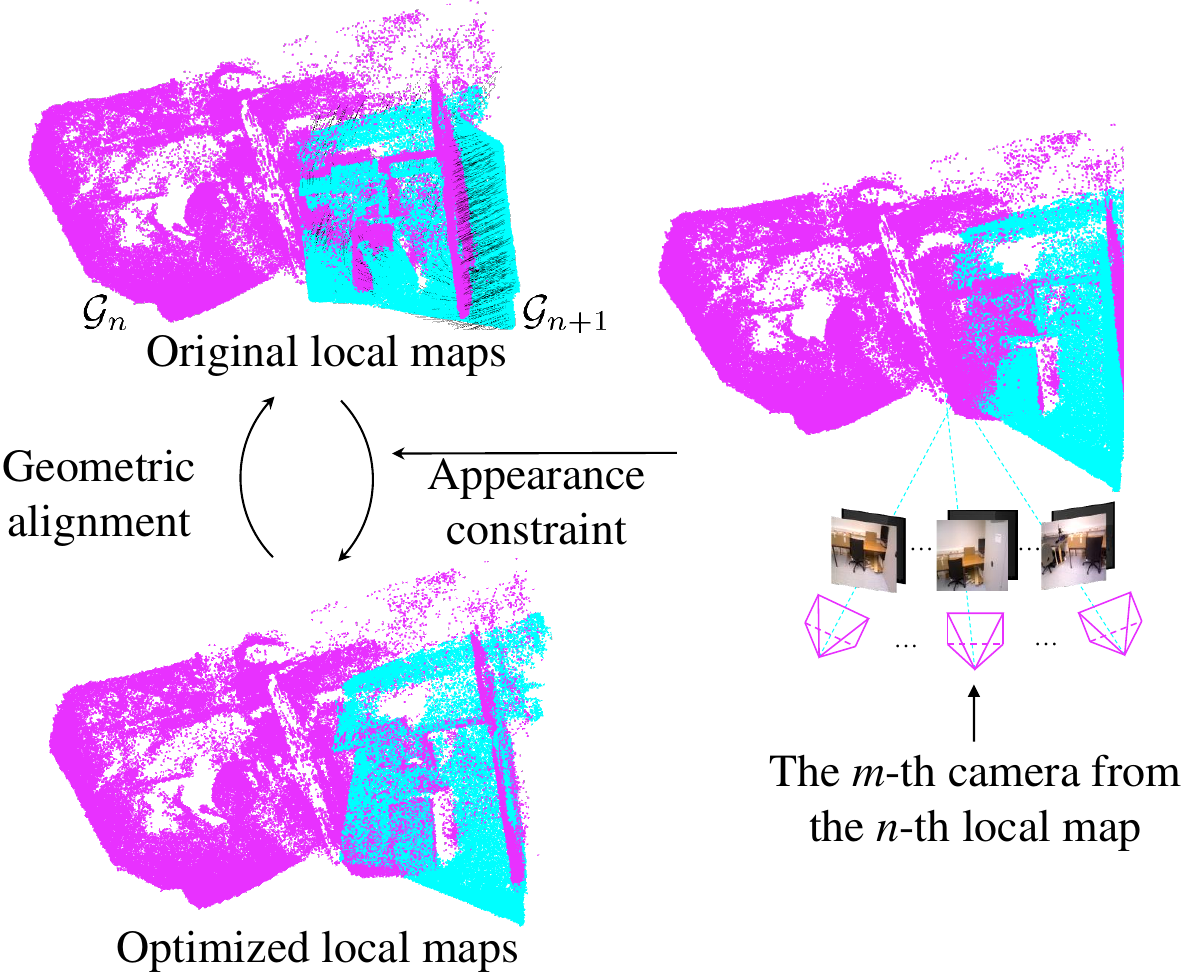}
	\caption{Optimization between 
    $n$-th and $(n+1)$-th
    local maps. Here, we show the centers of Gaussians.  We iteratively align Gaussians~$\mathcal{G}_{n+1}$ to Gaussians~$\mathcal{G}_{n}$ based on geometric constraint. To improve the robustness of optimization, we integrate the appearance constraint into each iteration. Gaussians~$\mathcal{G}_{n+1}$ are rendered by multiple cameras associated with Gaussians~$\mathcal{G}_{n}$.}
	\label{fig:submap}
\end{figure}
In the above, we generate and optimize each local map independently.
Neighboring local maps lack effective constraints to connect them. Accordingly, these local maps may exhibit appearance inconsistency and geometric non-alignment.
To solve this problem, we leverage the co-visibility constraint between neighboring local maps.
For ease of understanding, let us first consider the $n$-th and $(n+1)$-th local maps
for illustration.

For each local map, we use the centers of Gaussians to generate a 3D point cloud in the world frame.
Given two neighboring point clouds that partly 
deviate from each other due to noise,
a straightforward strategy is to use the iterative closest points (ICP) algorithm~\cite{icp_pami} to align them. However, in practice, the deviation between two point clouds may be relatively large.
The pure ICP-based strategy 
is prone to 
get stuck in a local optimum.
To align two local maps more robustly,
we integrate the appearance constraint
into ICP.      

    As shown in Fig.~\ref{fig:submap},
we first follow ICP to establish tentative 3D-3D point correspondences based on the shortest distance, and then use these correspondences to roughly estimate the relative transformation $\Delta \mathbf{T}_{n,n+1}$.
After that, instead of directly using $\Delta \mathbf{T}_{n,n+1}$ to update point correspondences, we exploit the 
appearance losses
to optimize $\Delta \mathbf{T}_{n,n+1}$. 
Specifically,  in the world frame,
we use $\Delta \mathbf{T}_{n,n+1}$ to transform the $(n+1)$-th local map~$\mathcal{G}_{n+1}^{\mathcal{W}}$. 
This map is further transformed to the $m$-th camera frame of the $n$-th local map  based on the
$m$-th camera pose
($1\leq m \leq M$).
Then we
 render these Gaussians into RGB and depth patches: $\tilde{I}_m(\Delta \mathbf{T}_{n,n+1}), \tilde{D}_m(\Delta \mathbf{T}_{n,n+1})=\pi[\mathcal{G}_{n+1}^{\mathcal{W}},\Delta \mathbf{T}_{n,n+1},\mathbf{T}_m]$.
 We use these rendered  patches and their corresponding observed patches
 to define the photometric and depth losses, optimizing the transformation~$\Delta \mathbf{T}_{n,n+1}$:
\begin{equation}
\begin{split}
     \underset{\Delta \mathbf{T}_{n,n+1}}{\min} & \sum_{m=1}^M 
\lambda_{\textnormal{P}} \cdot L_{\textnormal{P}}
\Big( \tilde{I}_{m}( \Delta \mathbf{T}_{n,n+1} ) \Big) + \\
     &  
     \sum_{m=1}^M  \lambda_{\textnormal{D}} \cdot  L_{\textnormal{D}}  \Big(  \tilde{D}_{m}( \Delta \mathbf{T}_{n,n+1} )  \Big).
\end{split}
\end{equation}
The optimized transformation~$\Delta \mathbf{T}_{n,n+1}$
is then fed back to the second round of iteration to establish more accurate 3D-3D point correspondences.
We repeat the above process until convergence.
The appearance and geometric losses complement each other, which can  help optimization converge more reliably.
Based on the optimized transformation~$\Delta \mathbf{T}_{n,n+1}$, we can better align neighboring local maps.

In addition, we can easily extend the above optimization method to loop closure.
Briefly, we can use an arbitrary loop detection algorithm to identify two local maps that partly overlap each other. Then our method can estimate the transformation between two local maps and use this transformation to perform trajectory correction.

\section{Camera Localization}
\label{sec:cam_localization}
In this section, we introduce how we localize the camera using not only static background, but also dynamic foreground.
Our method leverages both geometric and appearance constraints.
In the following, 
we consider the $t$-th and $(t+1)$-th RGB-D images for illustration.
We assume that both static background and dynamic foreground at time~$t$ have been reconstructed, and the camera pose~$\mathbf{T}_{t}$ has been estimated. We aim to estimate the camera pose at time~$t+1$.

\subsection{Two-stage Localization Strategy}
\label{subsec:two_stage_loc}
We propose a two-stage strategy to estimate the camera pose in a coarse-to-fine manner.
In the first stage, we only use the static background to obtain the initial estimation of the camera pose.
In the second stage, we simultaneously exploit the static background and dynamic foreground to refine the camera pose.
Note that we do not consider dynamic foreground in the first stage. The reason is that at time~$t+1$, we do not have a prior 3D map of dynamic foreground in the world frame which is consistent with the observed RGB-D images. By contrast, this 3D map can be roughly obtained
after the first-stage estimation
(see below), 
and thus can be used in the second stage.
Compared with the previous works~\cite{sun2018motion,yu2018ds,jiang2024rodyn} that only leverage static background, the above strategy can use more observations to compensate for noise and thus improve the accuracy of camera localization.
As to the constraints for optimization, we introduce not only the appearance constraints of Gaussians, but also the geometric constraints of optical flows.
In the following, we first consider the appearance constraints to illustrate the pipeline of our two-stage localization strategy. Then we will present the integration of geometric constraints in 
the next subsection.

In the first stage, given the background Gaussians~$\mathcal{B}_{t}^{\mathcal{W}}$ obtained at time~$t$ in the world frame (see Section~\ref{sec:mapping_background}), we use the unknown-but-sought camera pose~$\mathbf{T}_{t+1}$ to transform them to the $(t+1)$-th camera frame. We 
fix the attributes of the transformed Gaussians and render these Gaussians as RGB  and depth patches: $\tilde{I}_{t+1}(\mathbf{T}_{t+1}), \tilde{D}_{t+1}(\mathbf{T}_{t+1}) =
    \pi [\mathcal{B}_{t}^{\mathcal{W}},\mathbf{T}_{t+1}]$.
Then we use these rendered patches and their corresponding
observed RGB-D patches to define the photometric and
depth losses, optimizing the transformation~$\mathbf{T}_{k+1}$:
\begin{equation}
 \min_{\mathbf{T}_{t+1}} \	
  \lambda_{\textnormal{P}} \cdot
 L_{\textnormal{P}}
\Big( \tilde{I}_{t+1}(\mathbf{T}_{t+1}) \Big)
	  + 
   \lambda_{\textnormal{D}} \cdot  L_{\textnormal{D}}  \Big(  \tilde{D}_{t+1}(\mathbf{T}_{t+1})  \Big).
 \label{eq:first_stage}
\end{equation}
For optimization, we initialize the camera pose $\mathbf{T}_{t+1}$ based on the known camera pose~$\mathbf{T}_{t}$ 
and constant velocity motion model used in ORB-SLAM2~\cite{mur2017orb}.

In the second stage, we simultaneously consider the constraints of static background and dynamic foreground to refine the coarse camera pose~$\mathbf{T}_{t+1}$ estimated above.
The operation for static background is similar to that at the first stage. Therefore, we mainly introduce the usage of the dynamic foreground.
Recall that given RGB-D images~$(I_{t+1},D_{t+1})$ and coarse camera pose~$\mathbf{T}_{t+1}$ obtained in the first stage, we can roughly estimate a set of foreground Gaussians~$\mathcal{F}^{\mathcal{W}}_{t+1}$ 
 in the world frame (see Section~\ref{sec:mapping_dyn_foreground}).
Compared with background Gaussians optimized by multiple-view constraints,
foreground Gaussians 
may not be accurate enough due to fewer views,
and thus are treated as variables to optimize.
We use the coarse camera pose $\mathbf{T}_{t+1}$ 
to transform these Gaussians to the $(t+1)$-th camera frame, and render them as RGB and depth patches:
$\tilde{I}_{t+1}(\mathcal{F}^{\mathcal{W}}_{t+1},\mathbf{T}_{t+1}),\tilde{D}_{t+1}(\mathcal{F}^{\mathcal{W}}_{t+1},\mathbf{T}_{t+1}) = \pi[\mathcal{F}^{\mathcal{W}}_{t+1},\mathbf{T}_{t+1}]$.
We use these rendered patches and corresponding observed RGB-D patches to define the photometric and depth losses, optimizing Gaussians~$\mathcal{F}^{\mathcal{W}}_{t+1}$ and camera pose~$\mathbf{T}_{t+1}$:
\begin{equation}
	\begin{split}
  \min_{  \mathbf{T}_{t+1},\mathcal{F}^{\mathcal{W}}_{t+1} } &
  \lambda_{\textnormal{P}} \cdot
 L_{\textnormal{P}}
\Big( \tilde{I}_{t+1}(\mathcal{F}^{\mathcal{W}}_{t+1},\mathbf{T}_{t+1}) \Big)+
  \\ &
	\lambda_{\textnormal{D}} \cdot  L_{\textnormal{D}}  \Big(  \tilde{D}_{t+1}(\mathcal{F}^{\mathcal{W}}_{t+1},\mathbf{T}_{t+1})  \Big).
	\end{split}
    \label{eq:foreground_localization}
\end{equation}
Since the above strategy uses as much information as possible, it can effectively improve the accuracy of localization, as will be shown in the experiments. 

\subsection{Geometric Constraint of Optical Flows} 
\label{subsec:opt_flow}
The above appearance constraint-based localization can lead to accurate camera localization when the movements of cameras and dynamic objects are moderate. However, this strategy may be unreliable in the case of large movement.
The reason is that 
the rendered and observed pixels are prone to be wrongly associated, 
resulting in a local optimum of optimization.
To improve the robustness of camera localization, we further leverage the geometric constraints by associating 3D Gaussians and 2D optical flows.
We consider the optical flows of both static background (for the above first- and second-stage localization) and dynamic foreground (for the above second-stage localization). 

\subsubsection{Static Background}
\begin{figure}[!t]
	\centering
	\includegraphics[height=0.36\textwidth]{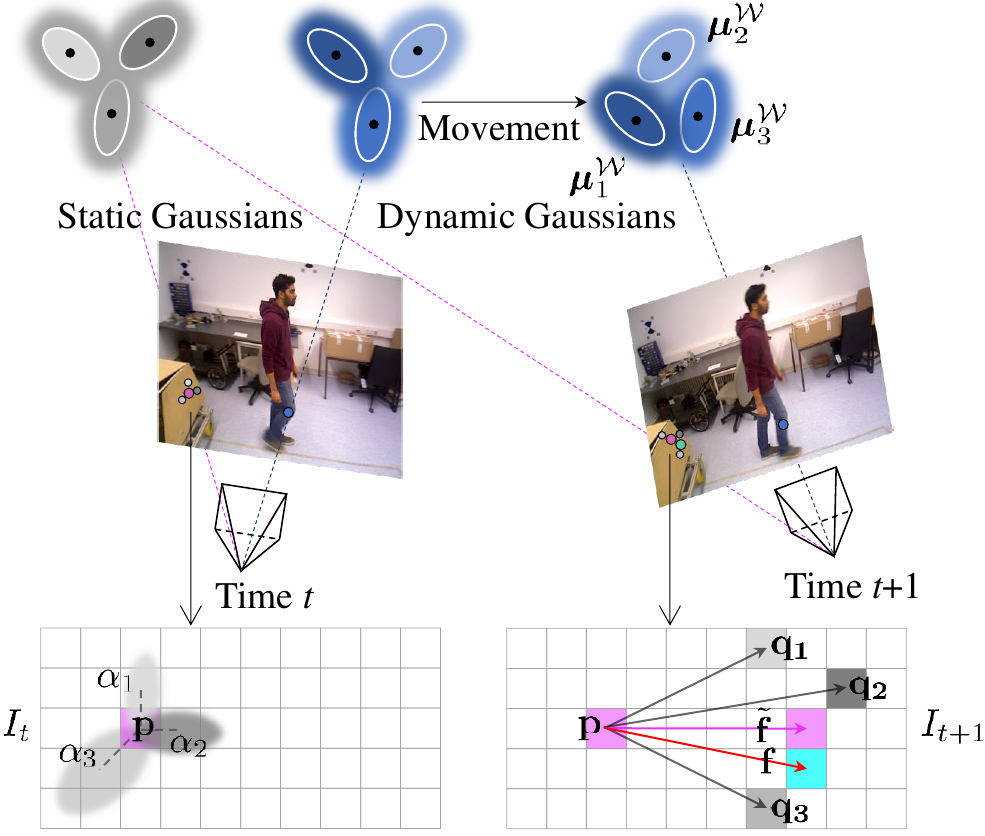}
    	\caption{Geometric constraint of optical flows. We mainly consider the static background for illustration. In the image~$I_{k+1}$, we compute the projected optical flow~$\tilde{\mathbf{f}}$ of the pixel~$\mathbf{p}$ based on the weighted combination of 2D  vectors $\mathbf{q}_k-\mathbf{p}$ (shown in black).  The projected optical flow~$\tilde{\mathbf{f}}$ should overlap the observed   optical flow~$\mathbf{f}$ estimated using the image pair $(I_k, I_{k+1})$.
        For dynamic foreground, the
generation of a projected optical flow additionally involves
the centers~$\{\boldsymbol{\mu}_k^{\mathcal{W}}\}$ of dynamic Gaussians at time $k+1$.
       }
	\label{fig:optical_flow}
\end{figure}

Optical flows of static background are caused by the camera movement, which provides a clue for camera pose estimation. 
First, we establish the connection between optical flows and background Gaussians. As shown in Fig.~\ref{fig:optical_flow}, a pixel~$\mathbf{p}$ in the image~$I_t$ corresponds to a set of 3D Gaussians.
The color of this pixel is determined by
a set of 2D Gaussians projected from these 3D Gaussians.
Each 2D Gaussian is associated with a  blending weight~$\alpha_k$ to encode its importance for rendering (see Section~\ref{label:subsub_pre_GS}).
We neglect 2D Gaussians whose weights are too small and assign the weights of the remaining $K$  2D Gaussians to their corresponding  3D Gaussians.
Then we transform 
these weighted 3D Gaussians from the world frame to the $(t+1)$-th camera frame using the unknown-but-sought camera pose~$\mathbf{T}_{t+1}$.
After that, we project the centers of these 3D Gaussians onto the $(t+1)$-th image, obtaining $K$  projected points~$\{\mathbf{q}_k (\mathbf{T}_{t+1}) \}_{k=1}^K$  with respect to the camera pose~$\mathbf{T}_{t+1}$.
The pixel~$\mathbf{p}$ and each projected point~$\mathbf{q}_k$ define a 2D vector by~$\mathbf{q}_k(\mathbf{T}_{t+1}) - \mathbf{p}$.
We compute the weighted sum of these vectors based on the blending weights $\alpha_k$ of Gaussians, generating the optical flow~$\tilde{\mathbf{f}}$ of the pixel~$\mathbf{p}$:
\begin{equation}
	\tilde{\mathbf{f}}(\mathbf{T}_{t+1}) = \sum_{k=1}^K\alpha_k \cdot \Big( \mathbf{q}_k(\mathbf{T}_{t+1}) - \mathbf{p} \Big).
 \label{eq:proj_flow}
\end{equation}
Since~$\tilde{\mathbf{f}}$ is generated by projecting 3D Gaussians, we call it the
 ``projected'' optical flow.
$\tilde{\mathbf{f}}$ can be regarded as a function with respect to the camera pose~$\mathbf{T}_{t+1}$.

The above generation of optical flow is in 3D. In the following, we compute the optical flow in 2D.
Given the RGB images~$I_t$ and $I_{t+1}$, we utilize the RAFT \cite{raft} algorithm to track pixels.
For the pixel~$\mathbf{p}$ in the image~$I_t$, we obtain its optical flow~$\mathbf{f}$ and call it the ``observed'' optical flow.
Ideally, the projected and observed flows should overlap each other. Based on this constraint, we minimize the difference between $\tilde{\mathbf{f}}(\mathbf{T}_{t+1})$ in Eq.~(\ref{eq:proj_flow}) and $\mathbf{f}$ to optimize the camera pose~$\mathbf{T}_{t+1}$. 
In practice, we establish $J$ pairs of projected and observed optical flows~$\{ ( \tilde{\mathbf{f}}_j
, \mathbf{f}_j ) \}_{j=1}^J$,
formulating the camera pose optimization by
\begin{equation}
\min_{\mathbf{T}_{t+1}}
\sum_{j=1}^J
  \| \tilde{\mathbf{f}}_j(\mathbf{T}_{t+1})  - \mathbf{f}_j \|_2.
\end{equation}

\subsubsection{Dynamic Foreground}
Optical flows of dynamic foreground are caused by not only the camera movement, but also  the motion of dynamic objects,
compared with the above static foreground.
As shown in Fig.~\ref{fig:optical_flow}, each  point~$\mathbf{q}_k$ projected from a dynamic Gaussian is with respect to  the camera pose~$\mathbf{T}_{t+1}$, as well as the Gaussian center~$\boldsymbol{\mu}_k^{\mathcal{W}}$ in the world frame at time~$t+1$.
Accordingly, we extend the computation of the projected optical flow of static background (see Eq.~(\ref{eq:proj_flow})) into
\begin{equation}
    	\tilde{\mathbf{f}}(\mathbf{T}_{t+1}, \{ \boldsymbol{\mu}_{k}^{\mathcal{W}} \}_{k=1}^K  ) = \sum_{k=1}^K\alpha_k \cdot \Big( \mathbf{q}_k(\mathbf{T}_{t+1}, \boldsymbol{\mu}_{k}^{\mathcal{W}}) - \mathbf{p} \Big).
        \label{eq:opt_flow_dyn}
\end{equation}
The projected optical flow~$\tilde{\mathbf{f}}$ is additionally with respect to the Gaussian center~$\boldsymbol{\mu}_k^{\mathcal{W}}$, compared with Eq.~(\ref{eq:proj_flow}).
Then by analogy with the static foreground, we minimize the difference between the projected flow~$\tilde{\mathbf{f}}$ in Eq.~(\ref{eq:opt_flow_dyn}) and  the observed flow~$\mathbf{f}$ extracted by RAFT algorithm. This achieves a joint optimization of the transformation~$\mathbf{T}_{t+1}$ and Gaussian centers~$\{\boldsymbol{\mu}_k^{\mathcal{W}}\}_{k=1}^K$.

\section{Experiments}
In this section, we first introduce the experimental setup, and then compare our method with state-of-the-art approaches. After that, we conduct ablation study to validate the effectiveness of the proposed modules.

\subsection{Experimental Setup}
\begin{figure}[!h]
\footnotesize
	\renewcommand\arraystretch{1.2}
	\renewcommand{\tabcolsep}{1pt} 
	\centering
\begin{tabular}{cc}
\includegraphics[width=0.49\linewidth]{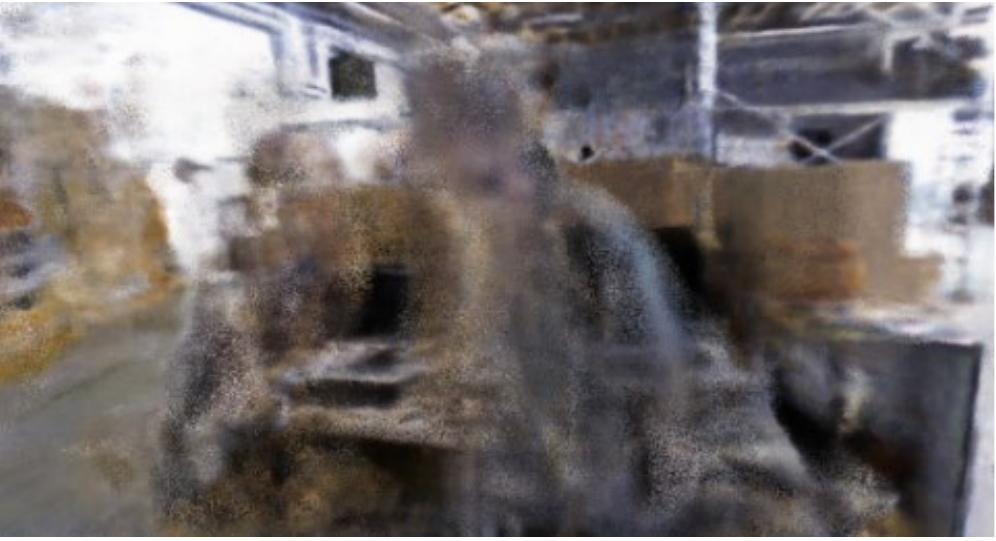} & 
\includegraphics[width=0.49\linewidth]{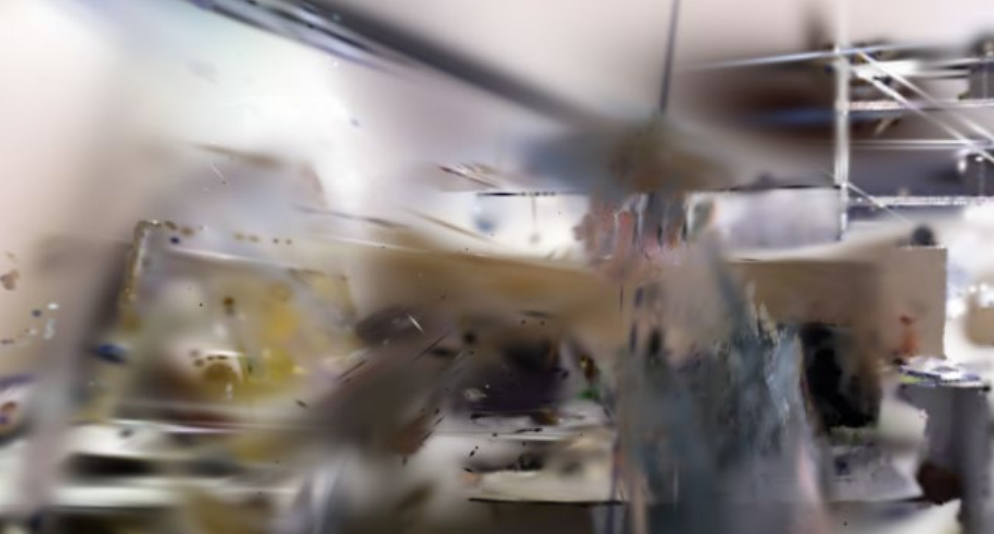} \\
(a) \textsf{ESLAM}~\cite{johari2023eslam} &
(b) \textsf{MonoGS}~\cite{GS-SLAM}  \\[0.5em]
\multicolumn{2}{c}{\includegraphics[width=0.83\linewidth]{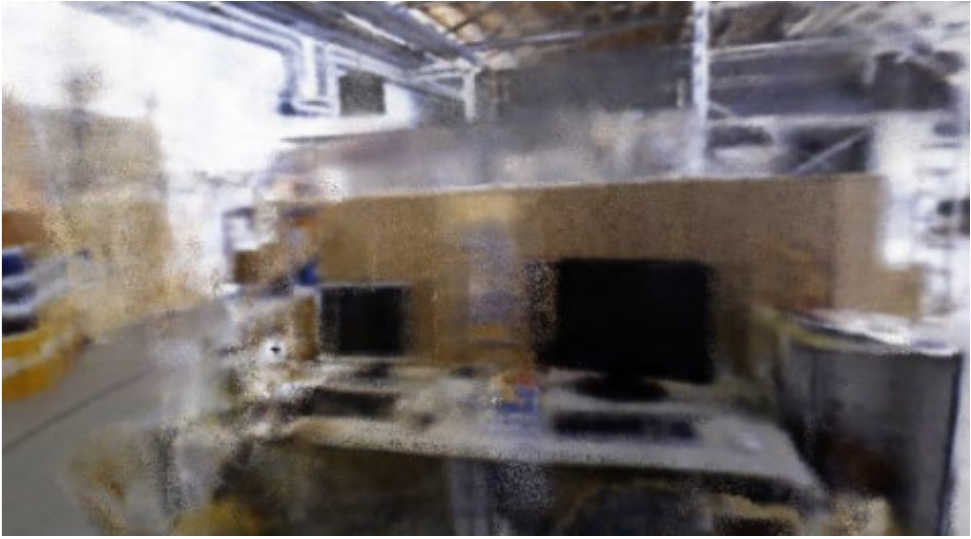}} \\
\multicolumn{2}{c}{(c) \textsf{Rodyn-SLAM}~\cite{jiang2024rodyn} } \\[0.5em]
\multicolumn{2}{c}{\includegraphics[width=0.83\linewidth]{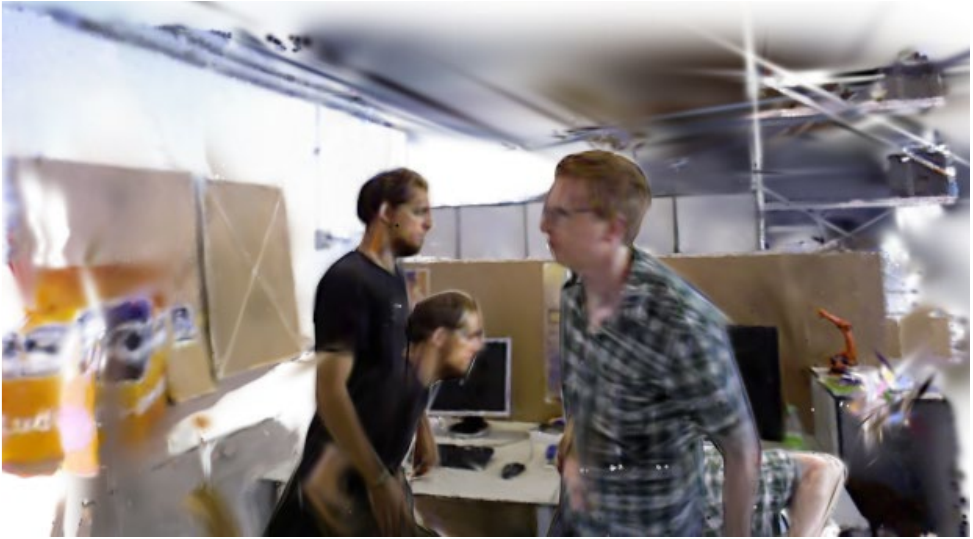}} \\
\multicolumn{2}{c}{(d) \textsf{PG-SLAM} (our) }
\end{tabular}
\caption{Representative environment mapping comparison between various SLAM methods on Sequence \texttt{walking\_rpy} of \textsf{TUM} dataset~\cite{tumbenchmark}. (a) \textsf{ESLAM} and (b) \textsf{MonoGS}~\cite{GS-SLAM} are originally designed for static environments, and cannot handle this dynamic scene well. 
(c) \textsf{Rodyn-SLAM} only focuses on mapping static background.
(d) Our \textsf{PG-SLAM} can reconstruct both static background and  dynamic humans at different times.}
       	\label{Tum_mapping}
\end{figure}
\begin{figure}[!t]
\footnotesize
\centering
\begin{tabular}{c}
		\includegraphics[width=0.9\linewidth]{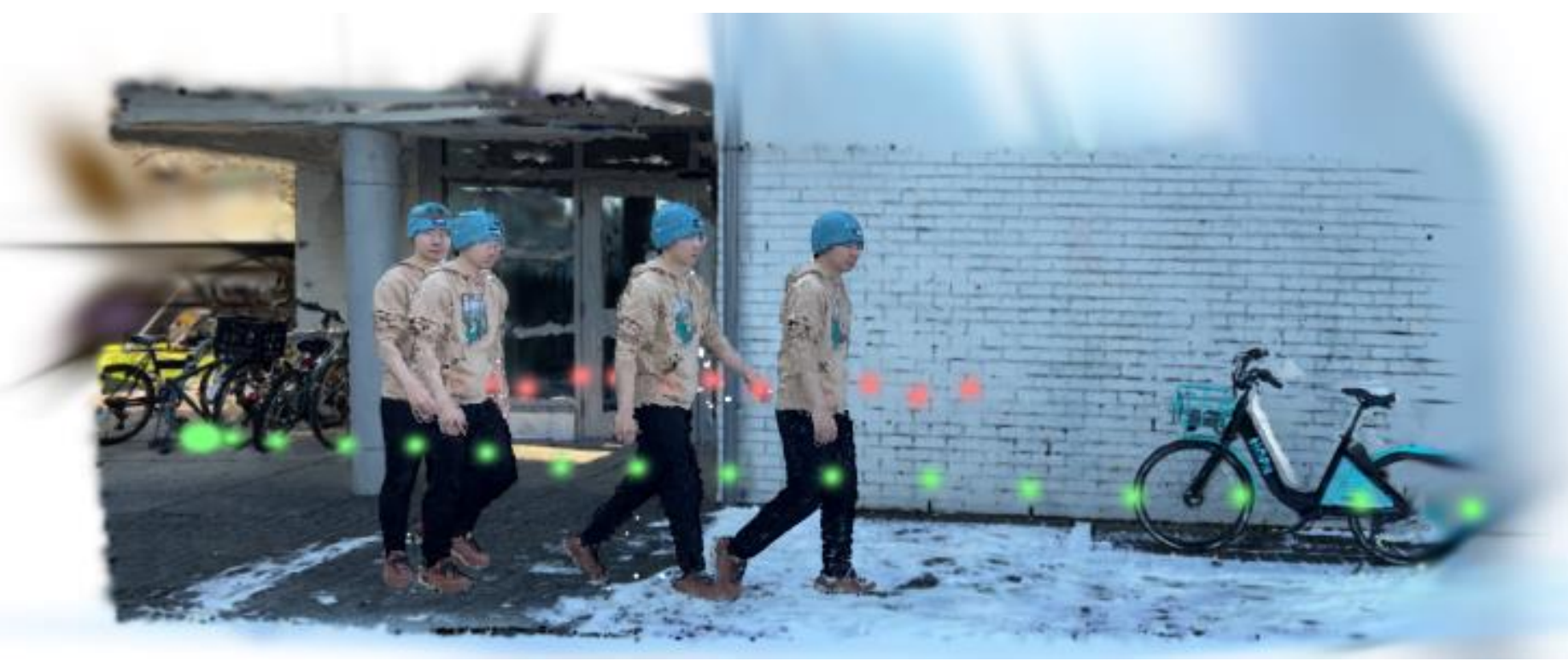} \\[0em]
  (a)  \\[0.5em]
		\includegraphics[width=0.9\linewidth]{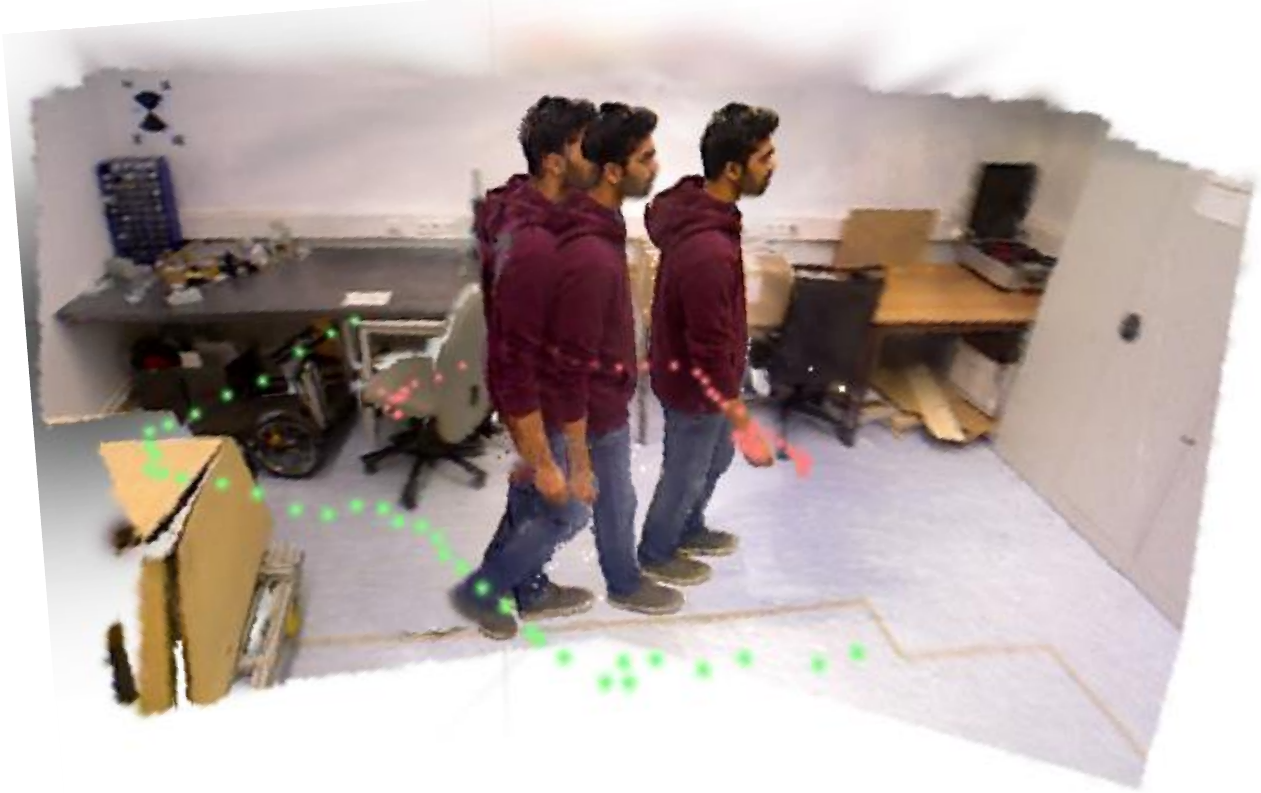}\\[0em]
		  (b)  
	\end{tabular}\\
       	\caption{Representative environment mapping results of our \textsf{PG-SLAM} on (a) Sequence~\texttt{bike} of \textsf{NeuMan} dataset~\cite{neuman}, and (b) Sequence~\texttt{moving\_box} of \textsf{TUM} dataset~\cite{tumbenchmark}. Our method can reconstruct not only static background, but also non-rigid human and rigid box at different times. The red and green dotted lines denote the trajectories of human and camera, respectively.}
       	\label{fig_3d_map}
\end{figure}

\begin{figure}[!t]
\footnotesize
\centering
\begin{tabular}{c}
		\includegraphics[height=0.48\linewidth]{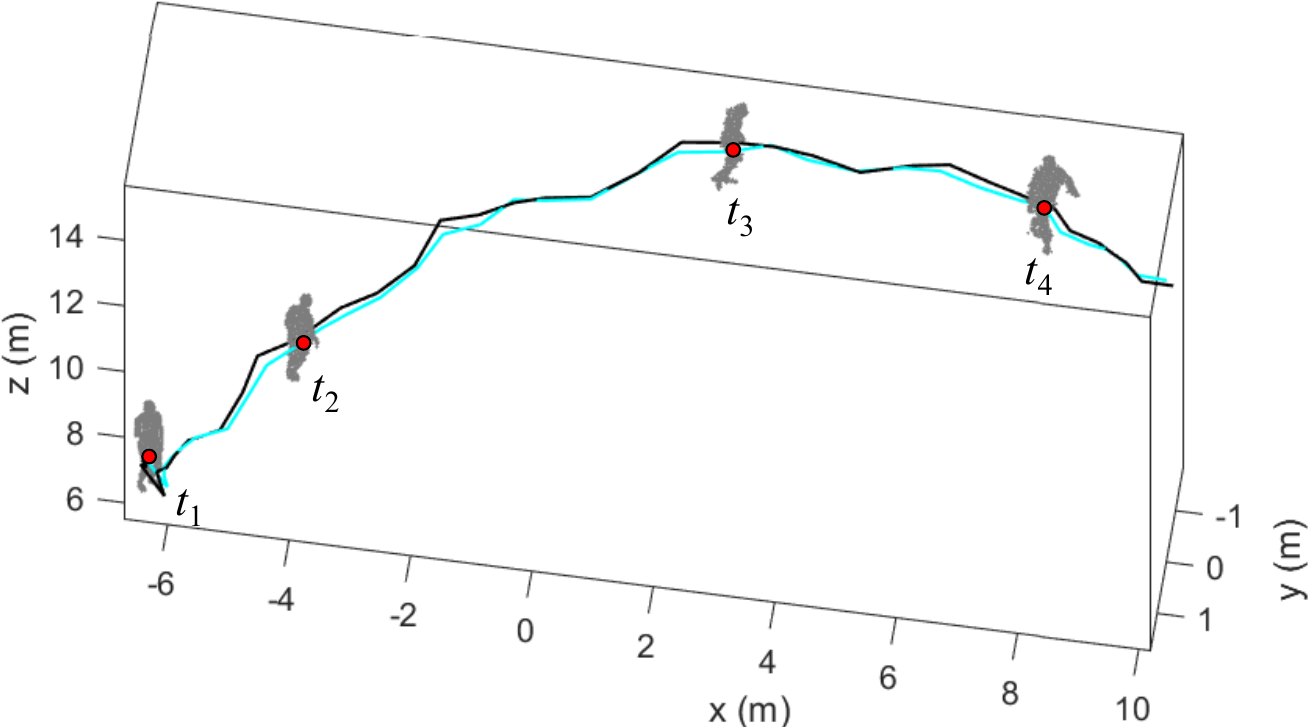} \\[0em]
  (a)    \\[0.5em]
		\includegraphics[height=0.3\linewidth]{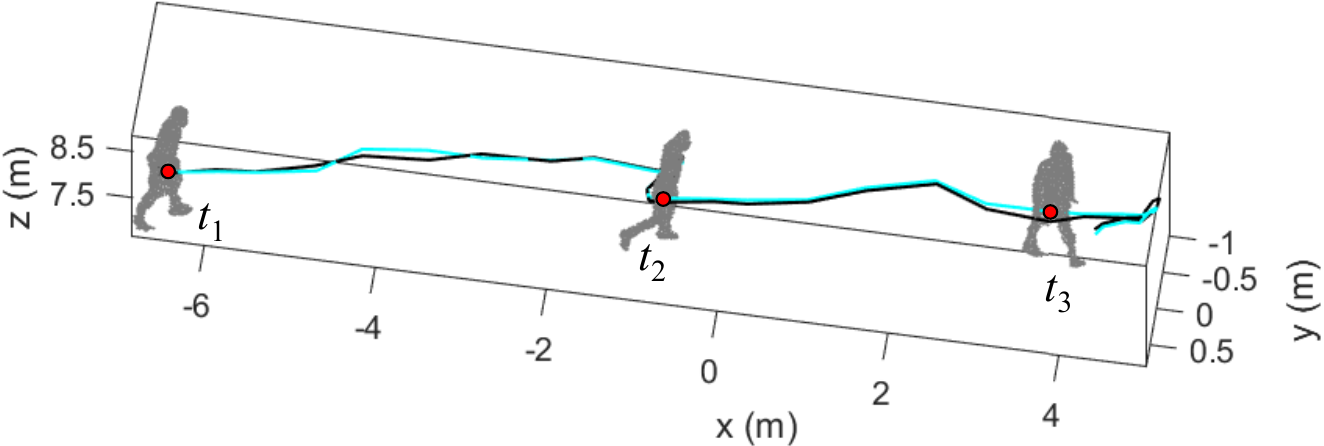}\\[0em]
		  (b)   
	\end{tabular}\\
       	\caption{Representative human trajectories estimated by our \textsf{PG-SLAM} on (a) Sequence \texttt{parking\_lot} and (b) Sequence \texttt{seattle} of \textsf{NeuMan} dataset~\cite{neuman}. The cyan and black lines denote the estimated and ground truth human trajectories, respectively. The centers of human Gaussians and the root joints of a human at some randomly selected times (such as $t_1, t_2, \cdots$) are shown in gray and red points, respectively.} 
       	\label{fig:human_traj}
\end{figure}
\subsubsection{Datasets}
We follow dynamic SLAM methods~\cite{jiang2024rodyn,bescos2021dynaslam,yu2018ds,zhang2020flowfusion} to conduct experiments on Bonn RGB-D Dynamic dataset~\cite{ReFusion} and TUM RGB-D dataset~\cite{tumbenchmark}. Moreover, we consider NeuMan dataset~\cite{neuman} to additionally evaluate the accuracy of our human localization.
For writing simplification, we denote the above datasets by \textsf{Bonn}, \textsf{TUM}, and \textsf{NeuMan} datasets, respectively.
We provide basic information as follows.
\begin{itemize}
    \item \textsf{Bonn} dataset was obtained in indoor environments. It includes several sequences with one or more humans performing various actions and items exhibiting rigid transformations. 
    It provides ground-truth camera trajectories to
    evaluate the accuracy of camera localization.
    \item \textsf{TUM} dataset was established in indoor environments that involve both dynamic and static scenes. We use the sequences of dynamic scenes for evaluation.
    Similar to the above \textsf{Bonn} dataset, some sequences 
    include multiple dynamic humans. 
    This dataset also provides ground truth 
    camera trajectories for evaluation of camera localization.
    \item \textsf{NeuMan} dataset 
    was obtained in outdoor environments. 
    The quality of the depth image is partly affected by the limited measurement distance of the depth camera.
    Each sequence contains one human with various poses.
    Different from the above datasets, this dataset not only offers ground-truth trajectories of cameras, but also the reference positions of dynamic humans. 
\end{itemize}
  
\subsubsection{Evaluation Metrics}
As to 3D mapping, we mainly adopt the qualitative evaluation
since some datasets lack ground truth 3D structures.
To evaluate the accuracy of camera localization and human tracking, we first connect the camera centers and root joints of a human at different times to generate trajectories of camera and human, respectively. Then we adopt the widely-used  absolute trajectory error~\cite{UZH_ATE,tumbenchmark}
to measure the difference between the estimated and ground truth trajectories.
We report both root mean square error (RMSE) and standard deviation (SD) of this error.
Unless otherwise specified, the unit of the reported values is centimeters.

\subsubsection{Implementation Details}
We treat Gaussian-SLAM~\cite{gaussianslam} as the baseline to develop our SLAM method.
For appearance constraint,
we set the weights of the photometric and depth losses $\lambda_{\textnormal{P}}$ and $\lambda_{\textnormal{D}}$ (see Sections~\ref{sec:mapping_dyn_foreground}, \ref{sec:mapping_background}, and \ref{sec:cam_localization}) to 0.6 and 0.4, respectively.
We minimize the loss based on Adam following~\cite{splatam,zhu2022nice}. As introduced in Section~\ref{subsubsec:human_update}, the network~$\mathcal{D}$ to model the human deformation is based on MLP. 
Its input is the concatenation of the positional encoding result and human pose variation. Eight layers map the input into a 256-dimensional feature vector. Then this feature is processed by two independent layers to predict the variation of position and rotation, respectively.
We conduct experiments on a computer equipped with a CPU of E3-1226 
and a GPU of RTX 4090.

\subsection{Comparison with State-of-the-art Approaches}

\subsubsection{Methods for Comparison}
In the following, we denote our photo-realistic and geometry-aware SLAM method by \textsf{PG-SLAM}.
We compare our method with state-of-the-art approaches introduced in Section~\ref{sec:rel_work}:
\begin{itemize}
    \item \textsf{ESLAM} \cite{johari2023eslam}: The implicit representation-based method originally designed for static environments.
    It leverages the multi-scale axis-aligned feature planes.
    \item \textsf{MonoGS} \cite{GS-SLAM}: The GS-based approaches also targeted at static environments. It introduces several regularization strategies such as the isotropoic loss.
    \item \textsf{Rodyn-SLAM} \cite{jiang2024rodyn}: The implicit representation-based method suitable for dynamic environments. It directly eliminates dynamic foreground for robust estimation.
\end{itemize}
All the above methods are the differentiable rendering-based, which is helpful for a fair and unbiased comparison.

\begin{table*}[!t]
	\renewcommand\arraystretch{1.3} 
	\centering
\caption{Camera localization comparisons between various SLAM methods on 
three datasets.
We report the absolute trajectory error of camera trajectory.
}
        \begin{threeparttable}
\begin{tabular}{c}
\renewcommand{\tabcolsep}{6.33pt}
\renewcommand\arraystretch{1.3}
\begin{tabular}{c|cc|cc|cc|cc|cc|cc||cc}
        \toprule
        \multicolumn{15}{c}{\textsf{Bonn} dataset~\cite{ReFusion}}\\
        \midrule
        \multicolumn{1}{c}{Sequences}  & \multicolumn{2}{c}{\texttt{balloon}} & \multicolumn{2}{c}{\texttt{balloon2}} & \multicolumn{2}{c}{\texttt{ps\_track}} & \multicolumn{2}{c}{\texttt{ps\_track2}} & \multicolumn{2}{c}{\texttt{mv\_box}} & \multicolumn{2}{c}{\texttt{mv\_box2}}  & \multicolumn{2}{c}{Average} \\
        \midrule
        
        \textit{}  & RMSE & SD  & RMSE & SD  & RMSE & SD  &  RMSE & SD &  RMSE & SD   & RMSE & SD   & RMSE & SD\\
        \textsf{ESLAM}~\cite{johari2023eslam} &22.6 & 12.2  & 36.2 & 19.9 &48.0 & 18.7  & 51.4 & 23.2  & 8.4 & 3.5 & 17.7 & 7.5 & 31.4 & 14.7\\
        \textsf{MonoGS}~\cite{GS-SLAM}  & 49.5 & 20.3 & 42.5 & 20.9  & 100.3  & 50.7 & 114.3 & 52.8  & 8.9 & 3.7 & 24.1  & 11.2 & 66.1 & 31.1\\
        \textsf{RoDyn-SLAM}~\cite{jiang2024rodyn}  & 7.9 & 2.7 & 11.5 & 6.1  & 14.5  & 4.6 & 13.8 & 3.5  & 7.2 & 2.4 & 12.6  & 4.7 & 12.3 & 4.4\\
        \textsf{PG-SLAM} (our) & \textbf{6.4} & \textbf{2.2} &\textbf{7.3} &\textbf{3.4}  & \textbf{5.0}  & \textbf{1.9} & \textbf{8.5} & \textbf{2.8 }  & \textbf{4.6} & \textbf{1.3} & \textbf{7.0}  & \textbf{2.0} &\textbf{6.5} &\textbf{2.2}\\
        \midrule
        \midrule
    \end{tabular}
    \\

\renewcommand{\tabcolsep}{4.69pt}
\renewcommand\arraystretch{1.3}
\begin{tabular}{c|cc|cc|cc|cc|cc|cc|cc||cc}
\multicolumn{17}{c}{\textsf{TUM} dataset~\cite{tumbenchmark}} \\
\midrule

\multicolumn{1}{c}{Sequences} & \multicolumn{2}{c}{\texttt{f3/wk\_xyz}} & \multicolumn{2}{c}{\texttt{f3/wk\_hf}} & \multicolumn{2}{c}{\texttt{f3/wk\_st}} & \multicolumn{2}{c}{\texttt{f3/st\_hf}}   & \multicolumn{2}{c}{\texttt{f3/st\_rpy}}  & \multicolumn{2}{c}{\texttt{f3/st\_st}}  & \multicolumn{2}{c}{\texttt{f3/st\_xyz}} & \multicolumn{2}{c}{Average} \\
\midrule
\textit{}  & RMSE & SD & RMSE & SD  &  RMSE & SD   & RMSE & SD & RMSE & SD & RMSE & SD & RMSE & SD & RMSE & SD\\
\textsf{ESLAM}~\cite{johari2023eslam} &45.7 & 28.5 & 60.8 & 27.9 &93.6 & 20.7  & 5.3 & 3.3  & 8.4 & 5.7 & 0.9 & 0.5 & 5.4 & 3.9 & 31.4  & 12.9  \\
\textsf{MonoGS}~\cite{GS-SLAM} & 130.8 & 55.4 & 73.5 & 42.1 & 16.3  & 2.5 & 10.5 & 2.9 & 25.0 & 7.3 & 0.75 & 0.42 & 1.9 & 0.6 & 37.0 & 15.9 \\
\textsf{RoDyn-SLAM}~\cite{jiang2024rodyn} & 8.3 & 5.5 & \textbf{5.6} & \textbf{2.8} & 1.7  & 0.9 & 4.4 & 2.2 & 11.4 & 4.6 & 0.76 & 0.43 & 5.0 & 1.0 & 5.3 & 2.5 \\
\textsf{PG-SLAM} (our) & \textbf{6.8} & \textbf{2.9} & 11.7 & 4.4 & \textbf{1.4}  & \textbf{0.6} & \textbf{4.0} & \textbf{1.5} & \textbf{5.4} & \textbf{2.4} & \textbf{0.72} & \textbf{0.39} & \textbf{1.5} & \textbf{0.5} & \textbf{4.5} & \textbf{1.8}\\
            \midrule
            \midrule

\end{tabular}

\\
\renewcommand{\tabcolsep}{7.78pt}
\renewcommand\arraystretch{1.3}
\begin{tabular}{c|cc|cc|cc|cc|cc||cc}
            \multicolumn{13}{c}{\textsf{NeuMan} dataset~\cite{neuman}} \\
            \midrule
			\multicolumn{1}{c}{Sequences} & \multicolumn{2}{c}{\texttt{bike}} & \multicolumn{2}{c}{\texttt{citron}} & \multicolumn{2}{c}{\texttt{jogging}} & \multicolumn{2}{c}{\texttt{parkinglot}}  & \multicolumn{2}{c}{\texttt{seattle}}  & \multicolumn{2}{c}{Average} \\
			\midrule
			\textit{}  & RMSE & SD  & RMSE & SD   & RMSE & SD  &  RMSE & SD   & RMSE & SD   & RMSE & SD\\
			\textsf{ESLAM}~\cite{johari2023eslam}  &44.96 & 20.87  & 36.99 & 20.22 & 6.18 & 3.12  & 98.91 & 41.75  & 146.29 & 57.41& 66.66 & 28.67\\
			\textsf{MonoGS}~\cite{GS-SLAM}  &1.41 & 0.53  & 30.94 & 15.35 &16.47 & 6.52  & 3.03 & 1.42  & 0.86 & 0.34 & 10.52 & 4.83\\
			\textsf{RoDyn-SLAM}~\cite{jiang2024rodyn} & 1.38 & 0.54 & 4.62 & 3.51  & 1.02  & 0.46 & 3.07 & 1.20  & 2.65  & 1.02 & 2.54 & 1.35\\
			\textsf{PG-SLAM} (our)  & \textbf{1.15} & \textbf{0.47} &\textbf{2.43} &\textbf{0.92}  & \textbf{0.78}  & \textbf{0.45} & \textbf{2.39} & \textbf{1.12 }  & \textbf{0.53}  & \textbf{0.18} &\textbf{1.45} &\textbf{0.62}\\
\bottomrule  
		\end{tabular}
  \\
\end{tabular}
\end{threeparttable}
\label{table:com_traj}
\end{table*}

\subsubsection{Environment Mapping}
We compare different methods on multiple sequences of the above datasets. Fig.~\ref{Tum_mapping} shows a representative testing result on a sequence of \textsf{TUM} dataset, which involves both static background and two dynamic humans. For visual comparison, we render the same part of the scene from a novel view.
\textsf{ESLAM} and \textsf{MonoGS} do not distinguish between foreground and background. They inappropriately map the dynamic humans based on the strategies only suitable for static background,
leading to the fuzzy scene representation.
   \textsf{Rodyn-SLAM} filters out humans based on the estimated masks before mapping. While this method provides a satisfactory background reconstruction, it fails to model dynamic humans, and thus generates an incomplete map.
   Moreover, the loss of foreground information affects the accuracy of the map-based camera localization 
   and path planning.
   Our \textsf{PG-SLAM} can reconstruct not only the static background but also dynamic foreground.
   We merge the reconstructed background and foreground at different times, and render the merged elements at a certain time. Our method achieves a complete and photo-realistic scene representation.
   All the elements exhibit fine textures and details, as well as reasonable spatial relationship.

We further evaluate our~\textsf{PG-SLAM} on \textsf{Bonn} and \textsf{NeuMan} datasets. As shown in Fig. \ref{fig_3d_map}, each sequence contains a single dynamic human.
Similar to the above operations, on each sequence, we merge the reconstructed background and foreground at different times, and render the merged elements from a novel view at a certain time. We also present the estimated trajectories of humans and cameras.
Our~\textsf{PG-SLAM} can 
maintain the appearance consistency of both static background and dynamic foreground over time.

In addition, based on the ground truth human trajectories provided by \textsf{NeuMan} dataset, we evaluate the human localization of our~\textsf{PG-SLAM}.
We provide some qualitative results in Fig.~\ref{fig:human_traj} and also the quantitative results on all the sequences in the second part of Table~\ref{table:human_traj}.
Our estimated human trajectory is well aligned to the ground truth trajectory and the absolute trajectory error is small,
demonstrating high accuracy of our human localization.

\subsubsection{Camera Localization}

\begin{figure*}[!t]
	\footnotesize
	\centering
	\renewcommand{\tabcolsep}{3.0pt}
	\begin{tabular}{cc}
		$\textsf{MonoGS}$~\cite{GS-SLAM} 
		&
		\begin{tabular}{ccc}
			Sequence \texttt{Walking\_xyz} & Sequence \texttt{tracking\_person2} & Sequence \texttt{Citron}  
			\\
			\hline\\[-0.5em]
			\includegraphics[height=0.23\linewidth]{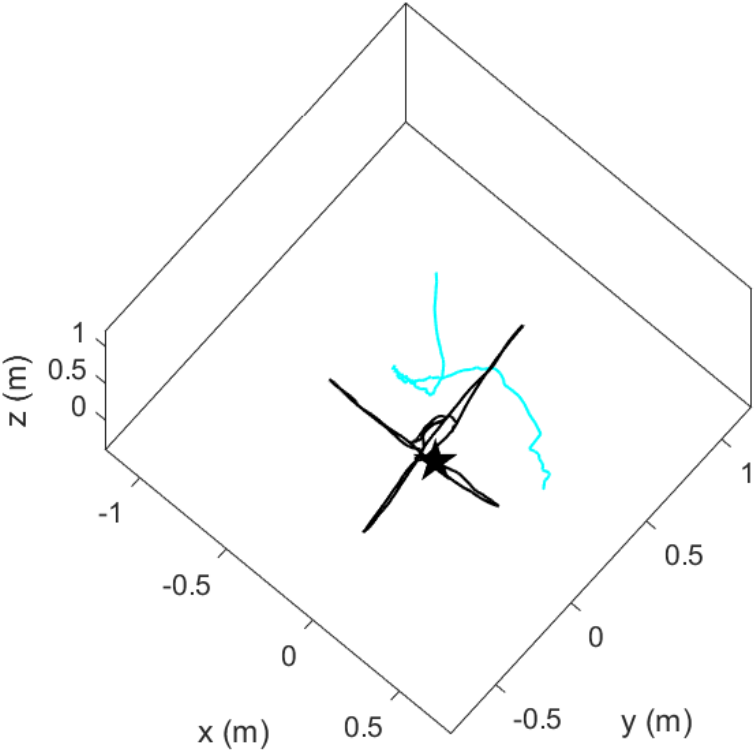} &
			\includegraphics[height=0.23\linewidth]{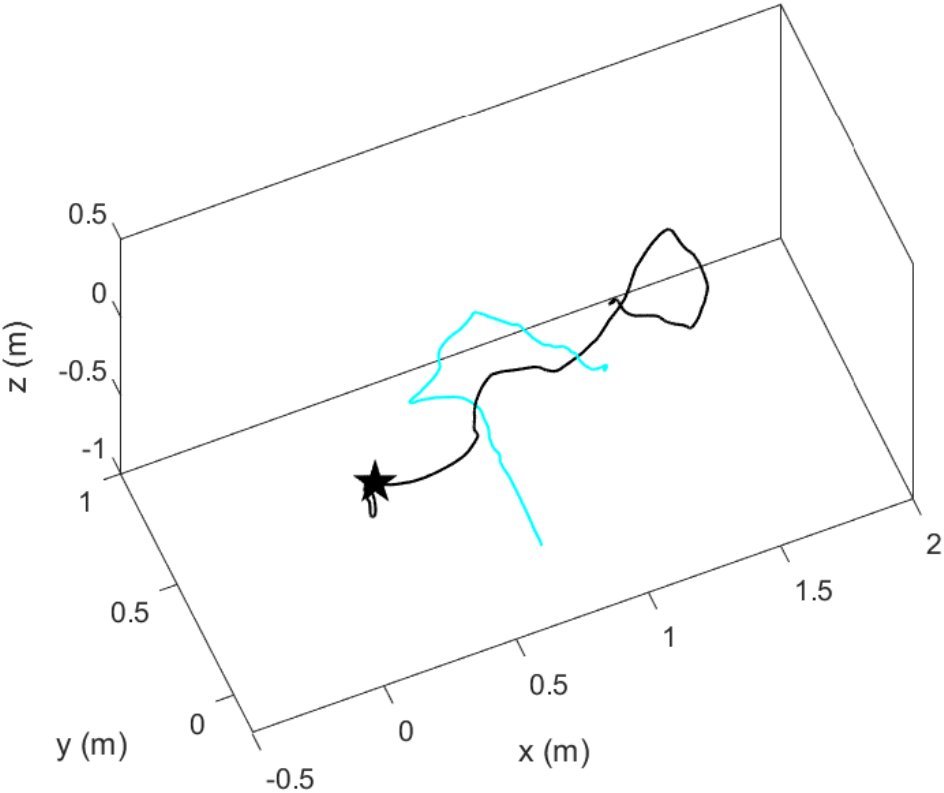} &

			\includegraphics[height=0.23\linewidth]{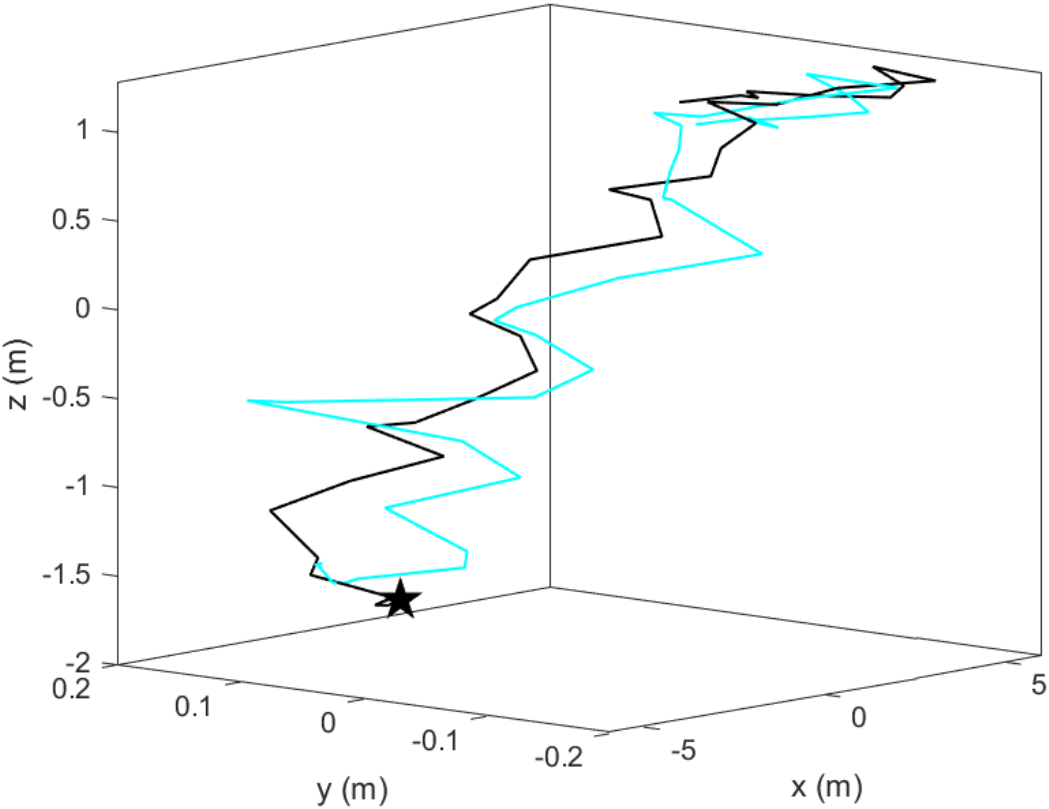}
			\\[0.3em]
		\end{tabular}
		\\
		$\textsf{ESLAM}$~\cite{johari2023eslam} 
		&
		\begin{tabular}{ccc}
			\includegraphics[height=0.23\linewidth]{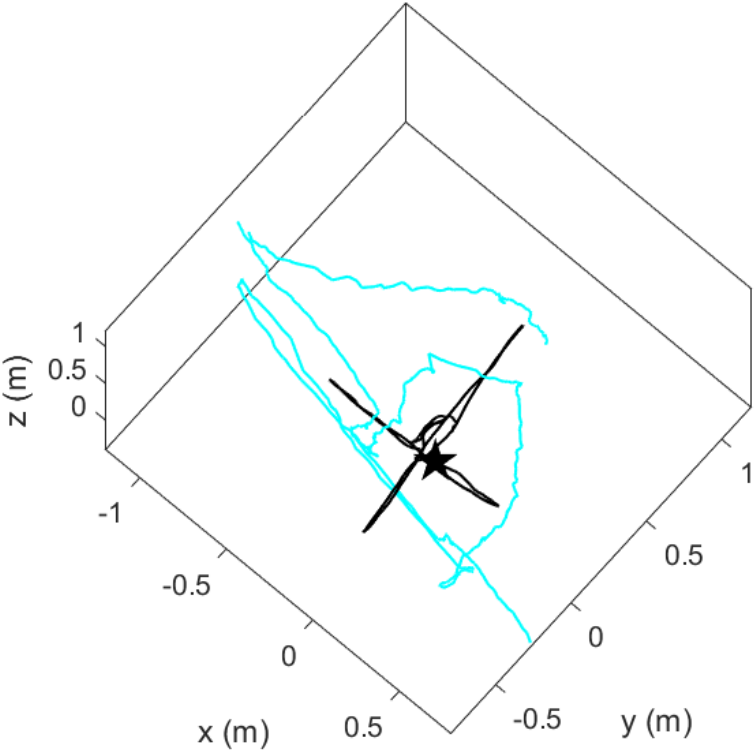} & 
			\includegraphics[height=0.23\linewidth]{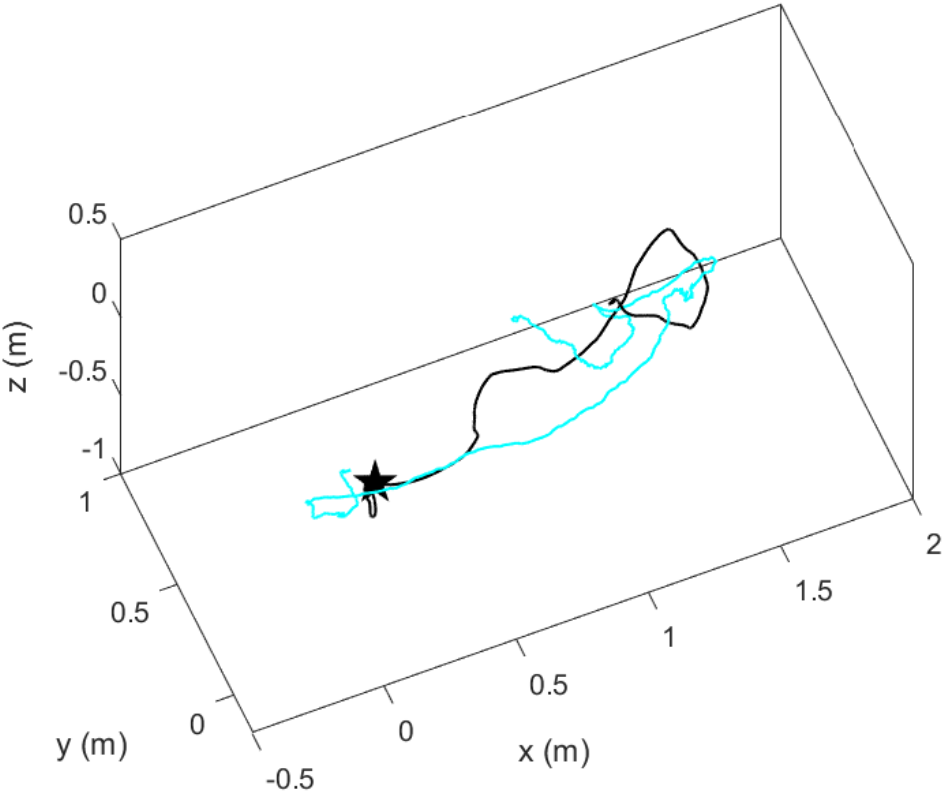} &
			\includegraphics[height=0.23\linewidth]{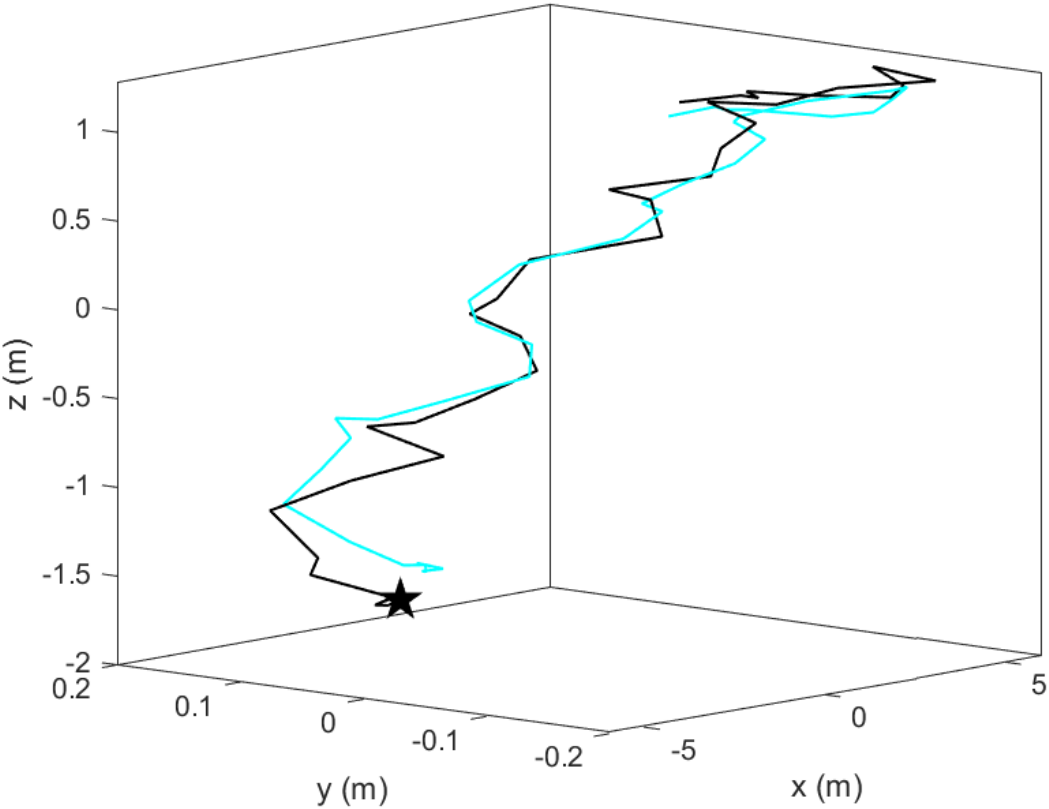}
			\\[0.3em]
		\end{tabular}
		\\
		$\textsf{Rodyn-SLAM}$~\cite{jiang2024rodyn}
		&
		\begin{tabular}{ccc}
			\includegraphics[height=0.23\linewidth]{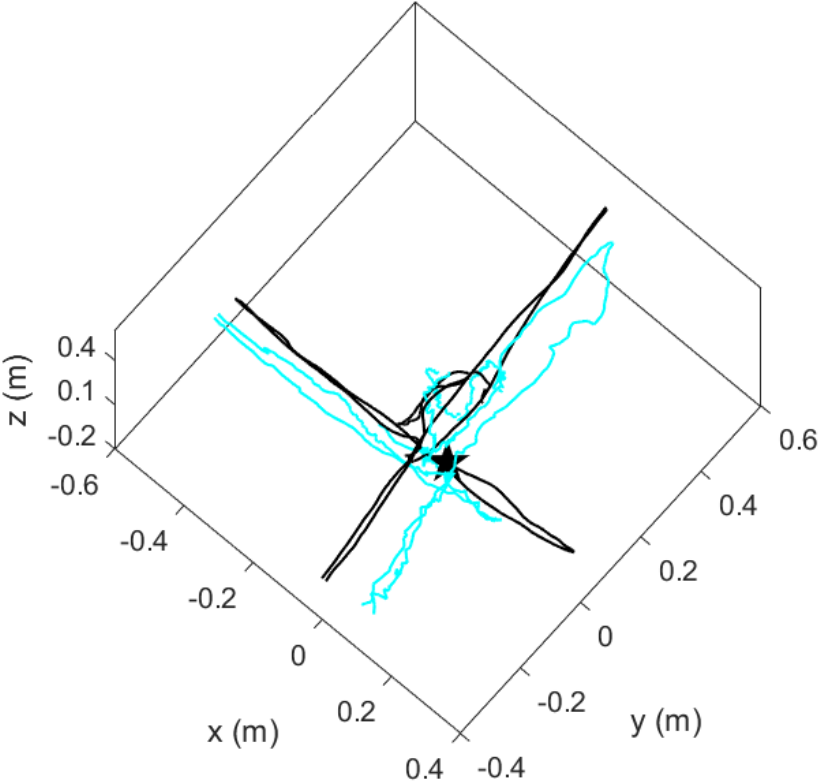} &
			\includegraphics[height=0.23\linewidth]{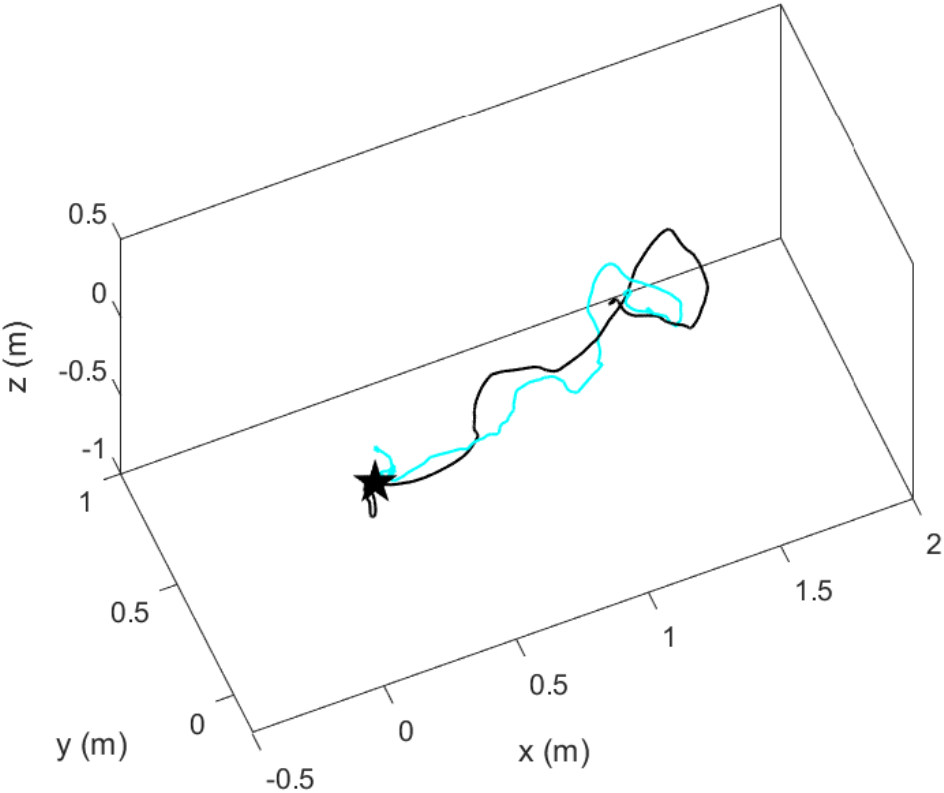} &
			\includegraphics[height=0.23\linewidth]{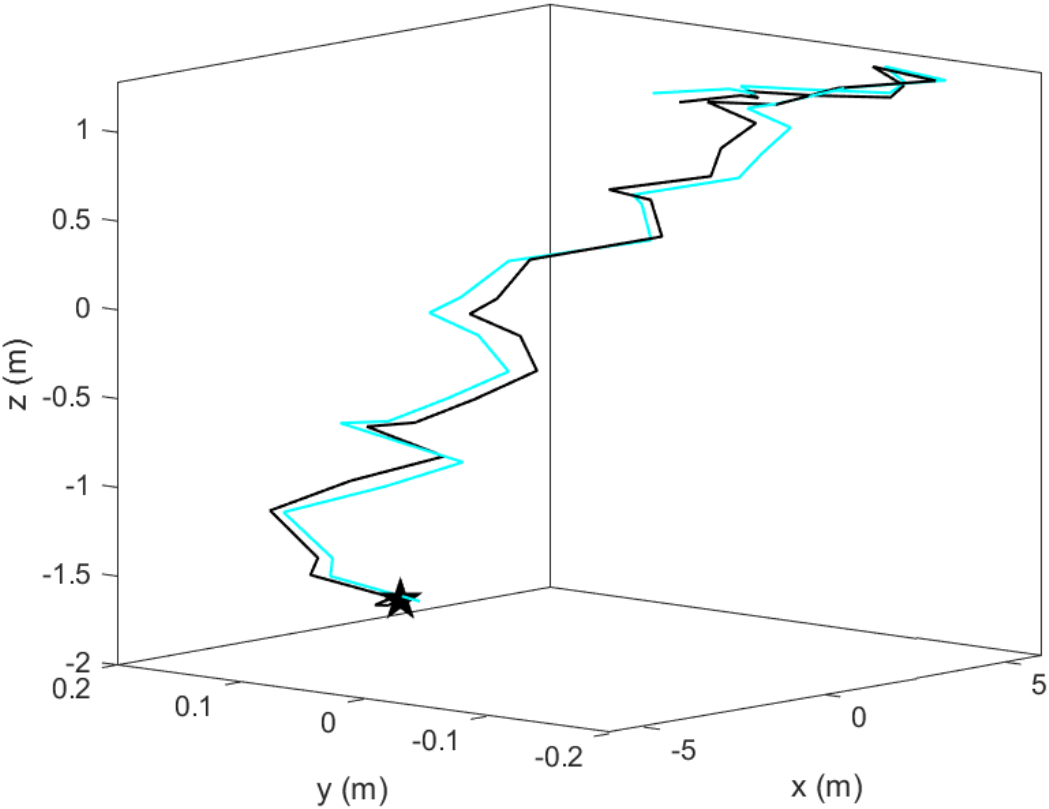}
			\\[0.3em]
		\end{tabular}
		\\ 
		\textsf{PG-SLAM} (our)
		&
		\begin{tabular}{ccc}
			\includegraphics[height=0.23\linewidth]{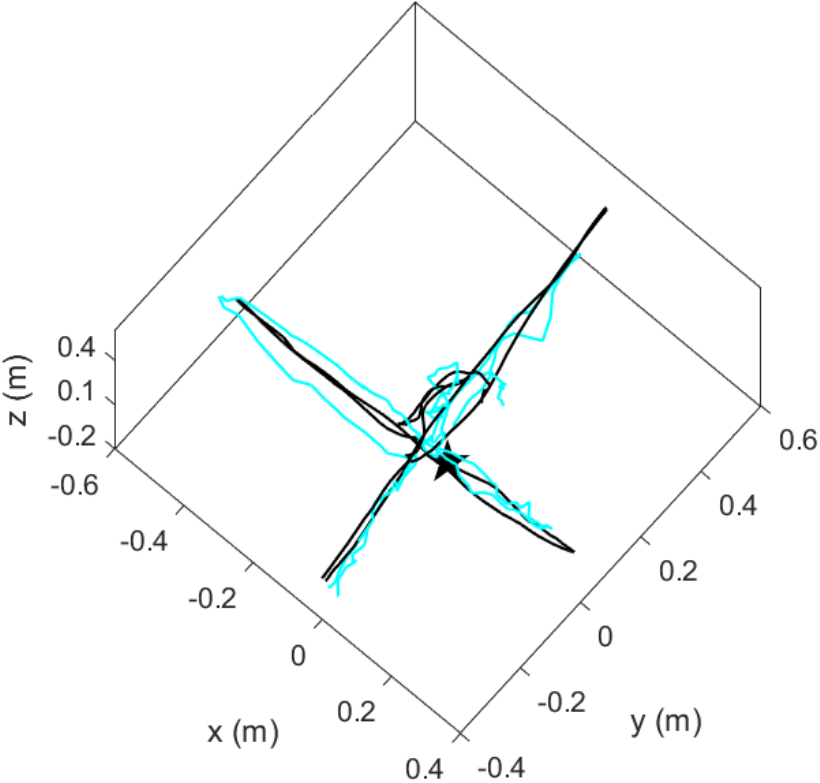} &
			\includegraphics[height=0.23\linewidth]{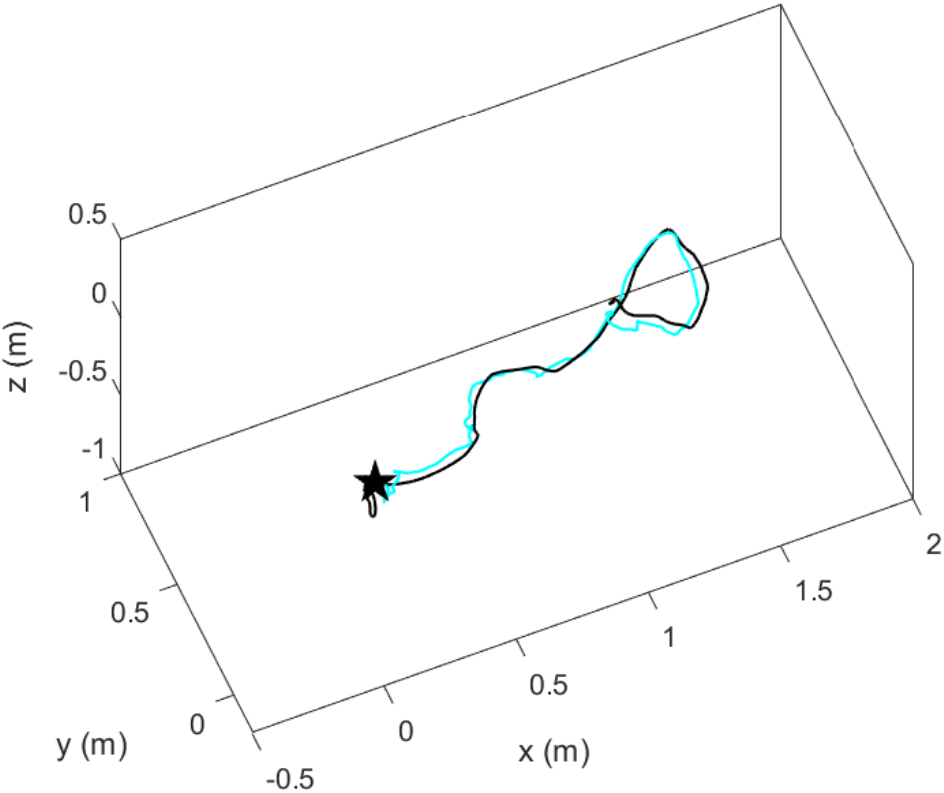} &
			\includegraphics[height=0.23\linewidth]{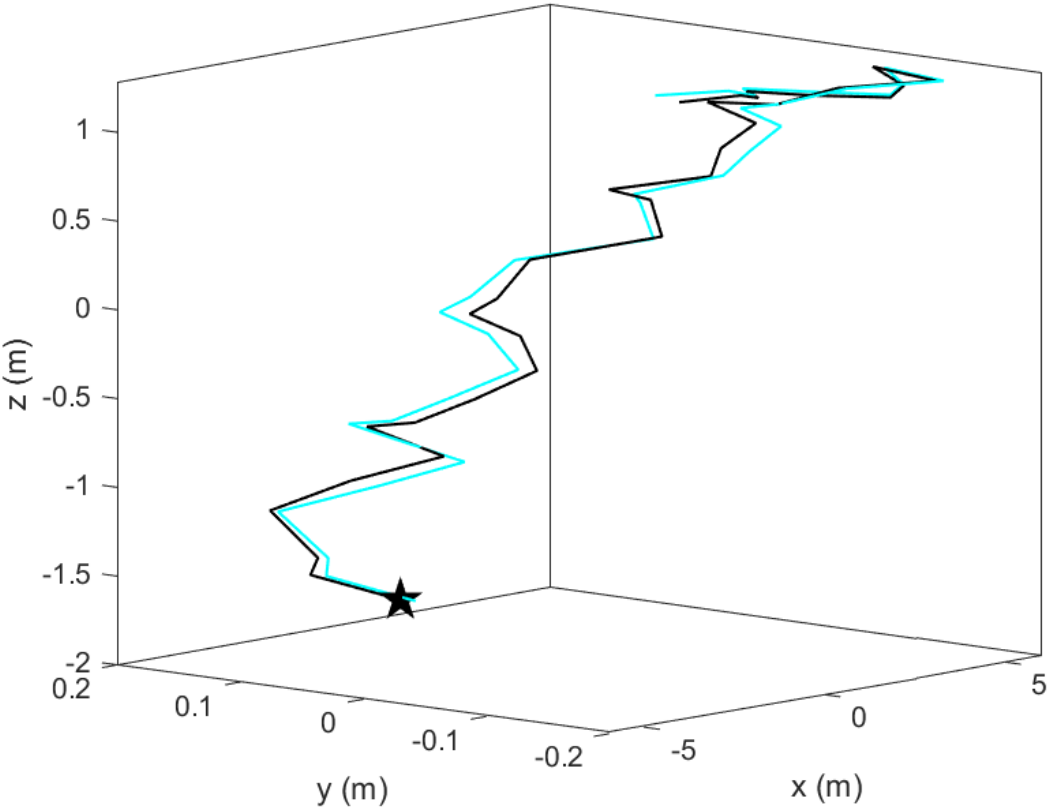}
			\\[0.3em]
		\end{tabular}
	\end{tabular}
	\\[-0.5em]
	\caption{Representative camera trajectories estimated by various SLAM methods on the Sequence \texttt{Walking\_xyz}       of \textsf{TUM} dataset~\cite{tumbenchmark}, Sequence \texttt{tracking\_person2} of \textsf{Bonn} dataset~\cite{ReFusion}, and Sequence \texttt{Citron} of \textsf{NeuMan} dataset~\cite{neuman}. The cyan and black lines denote the
		estimated and ground truth trajectories, respectively. Pentagram represents the starting point of camera trajectory.}
	\label{camera_traj}
\end{figure*}
\begin{figure*}[!h]
\footnotesize
\centering
\renewcommand{\tabcolsep}{5.0pt}
\begin{tabular}{ccc}
    \includegraphics[height=0.24\linewidth]{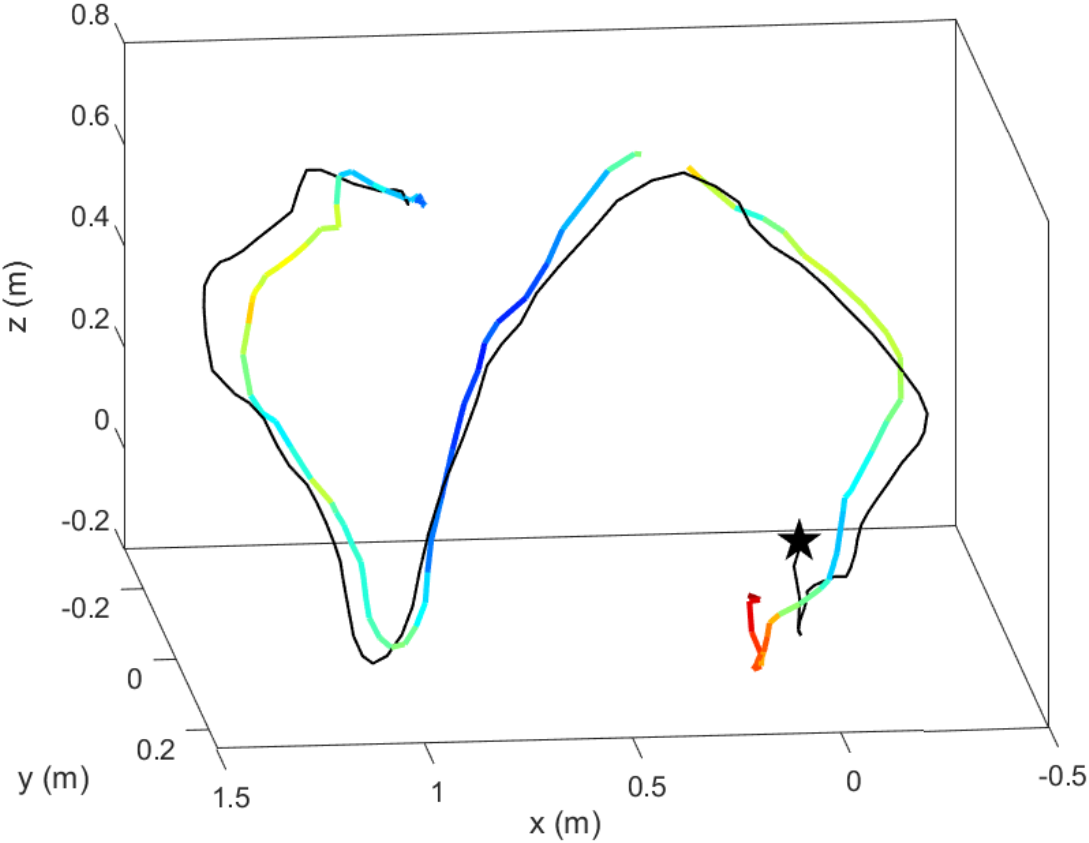} &
    \includegraphics[height=0.24\linewidth]{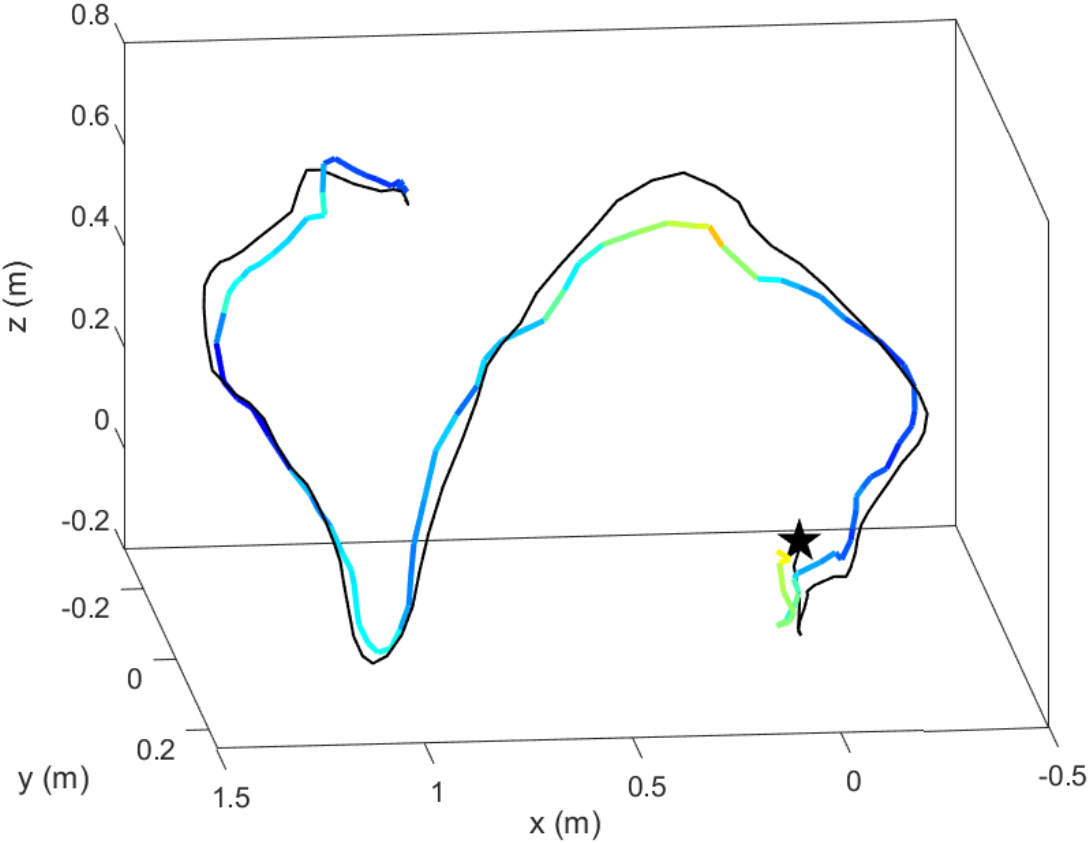} &
    \includegraphics[height=0.24\linewidth]{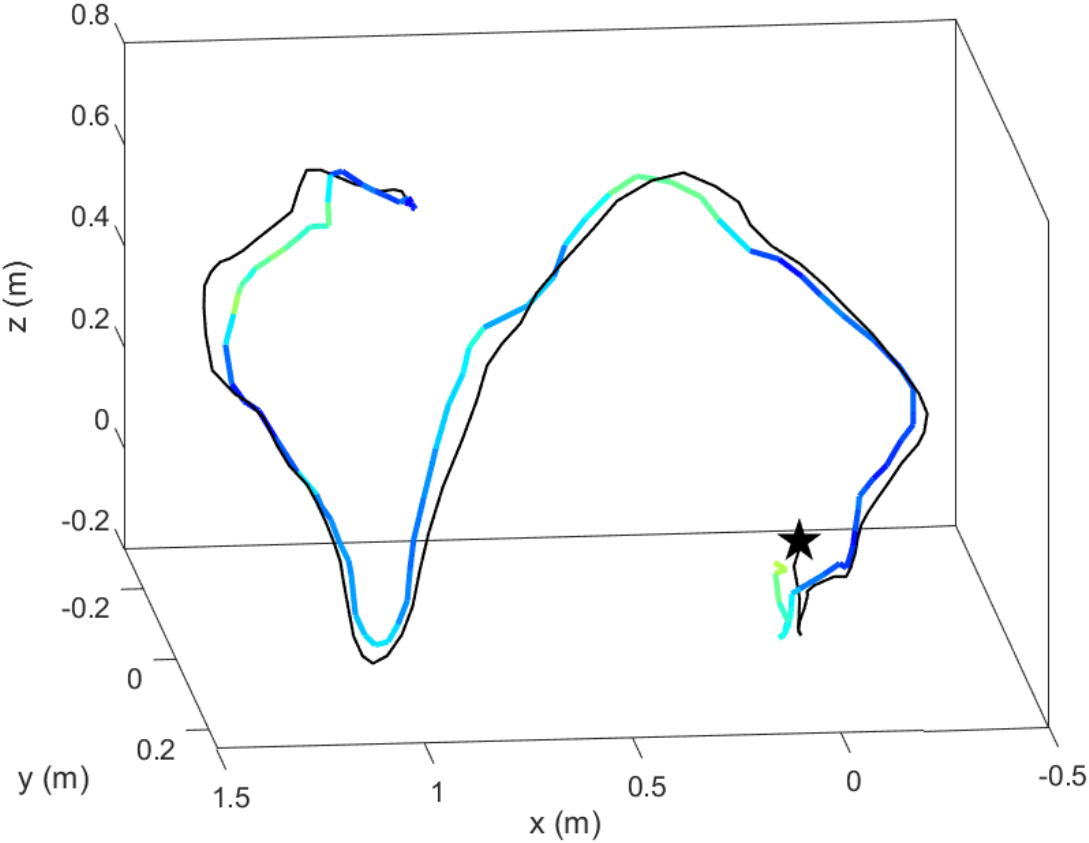} \\
    (a)	  & (b)   & (c)     
\end{tabular}
\\[0.5em]
\includegraphics[height=0.028\linewidth]{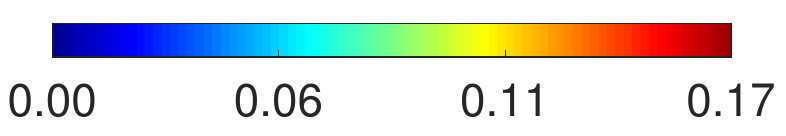}

\caption{Ablation study of camera localization using geometric constraint and dynamic foreground on Sequence \texttt{person\_track} of  \textsf{Bonn} dataset~\cite{ReFusion}. The colored and black lines denote the estimated and ground truth camera trajectories, respectively. Color bar indicates the magnitude of the absolute
trajectory error. Pentagram represents the starting point of camera trajectory.
(a) A version without using geometric constraint.
(b) A version without considering foreground. (c) Our complete method that simultaneously leverages the geometric and appearance constraints of both foreground and background.}

\label{ablation:module_traj}
\end{figure*}

We compare different methods on multiple sequences of \textsf{Bonn}, \textsf{TUM}, and   \textsf{NeuMan} datasets. 
We report quantitative results in Table~\ref{table:com_traj}, and also present some qualitative results on partial sequences in Fig. \ref{camera_traj}.
Overall, our \textsf{PG-SLAM} achieves state-of-the-art localization performance, demonstrating the effectiveness of our strategy that
simultaneously leverages geometric and appearance constraints of both foreground and background. 
In the following, we analyze the results on each dataset.

On \textsf{Bonn} dataset, the high proportion of non-rigid humans and rigid items (i.e., boxes and balloons) in the RGB-D images brings considerable challenges for camera localization. 
\textsf{ESLAM} and \textsf{MonoGS} are significantly affected, resulting in high trajectory errors.
\textsf{Rodyn-SLAM} overcomes the challenges to some extent, demonstrating the advantages of disentangling dynamic foreground from static background.
The accuracy of our \textsf{PG-SLAM}  significantly exceeds that of \textsf{Rodyn-SLAM} on all the sequences.

On \textsf{TUM} dataset, 
some sequences such as  \texttt{st\_st} and \texttt{st\_xyz} contain humans who sit on a chair and only exhibit slight motion variations. These cases approximate to the static environments, 
which allows \textsf{ESLAM} and \textsf{MonoGS} to achieve relatively good performance. 
However, both methods become unreliable on highly dynamic sequences such as \texttt{wk\_xyz} and \texttt{wk\_hf}. 
By contrast, \textsf{Rodyn-SLAM} can handle both cases more reliably by eliminating dynamic foreground.
Our \textsf{PG-SLAM} further improves the accuracy on most of the sequences.

On \textsf{NeuMan} dataset, dynamic humans account for a relatively small ratio of the RGB-D images.
Similar to the results on the above datasets, \textsf{ESLAM} and \textsf{MonoGS} perform unsatisfactorily, and \textsf{Rodyn-SLAM} partly improves the robustness. Our \textsf{PG-SLAM} achieves the best performance on all the sequences. 
In particular, the superiority of our \textsf{PG-SLAM} is significant on Sequence~\texttt{citron}
since human on this sequence exhibits a relatively large pose variation.

\begin{table}[!t]
    \renewcommand\arraystretch{1.3} 
    \renewcommand{\tabcolsep}{14pt} 
    \centering
    \caption{Ablation study of the geometric constraint-based localization on 
    three
    datasets. We report the absolute trajectory error of camera trajectory.}
    \begin{tabular}{c|cc|cc}
\toprule
    \multicolumn{1}{c}{Dataset}  &          
    \multicolumn{2}{c}{Without Constraint} & 
    \multicolumn{2}{c}{With Constraint}  
\\
\midrule
    & RMSE & SD  &  RMSE & SD
\\
    \textsf{Bonn}~\cite{ReFusion}  & 8.0 & 3.1 & \textbf{4.8}  &\textbf{1.6}
\\
    \textsf{NeuMan}~\cite{neuman}  & 1.3 & 0.4 & \textbf{0.78} &\textbf{0.4}
\\
    \textsf{TUM}~\cite{tumbenchmark}  & 8.6 & 3.5 & \textbf{6.8} &\textbf{2.9}
\\
\bottomrule 			
    \end{tabular}
\label{ablation:optical}
\end{table}

\subsection{Ablation Study}
\begin{table}[!t]
    \renewcommand\arraystretch{1.3} 
    \renewcommand{\tabcolsep}{13pt} 
    \centering
    \caption{Ablation study of localization using dynamic foreground on  
    three
    datasets. We report the absolute trajectory error of camera trajectory.}
	
    \begin{tabular}{c|cc|cc}
\toprule
    \multicolumn{1}{c}{Dataset}  &          
    \multicolumn{2}{c}{Without Foreground} & 
    \multicolumn{2}{c}{With Foreground}  
\\
\midrule
    & RMSE & SD  &  RMSE & SD
\\
    \textsf{Bonn}~\cite{ReFusion}  & 5.4 & 2.2 & \textbf{4.8}  &\textbf{1.6}
\\
    \textsf{NeuMan}~\cite{neuman}  & 0.96 & 0.4 & \textbf{0.78} &\textbf{0.4} 
\\
    \textsf{TUM}~\cite{tumbenchmark}  & 7.6 & 3.0 & \textbf{6.8} &\textbf{2.9}
\\
\bottomrule 			
    \end{tabular}
\label{ablation:foreground}
\end{table}

In this section, we conduct ablation studies of our proposed strategies and modules.

\subsubsection{Geometric Constraint-based Localization}
Recall that for camera localization, we additionally leverage the geometric constraint of optical flows to complement the appearance constraint (see Section~\ref{subsec:opt_flow}).
To validate its effectiveness,
we compare our complete method to the version without geometric constraint.
We report the quantitative results in Table~\ref{ablation:optical}
and provide a qualitative comparison in Figs.~\ref{ablation:module_traj}(a) and \ref{ablation:module_traj}(c).
On all the datasets, our geometric constraint can significantly improve the accuracy of camera localization,
especially on \textsf{Bonn} dataset where the proportion of dynamic objects is relatively high.

\subsubsection{Localization Using Dynamic Foreground}
\label{subsubsec:Ablation:Foreground}

Recall that our method 
leverages the dynamic foreground for camera localization (see Section~\ref{subsec:two_stage_loc}). 
We compare our complete method using both foreground and background with the version that only considers background.
As shown in Table~\ref{ablation:foreground}, by utilizing foreground information,  the accuracy of camera trajectory  can be  
effectively improved on all the datasets.
In particular, \textsf{Bonn} dataset contains sufficient observations of dynamic foreground including both non-rigid humans and rigid items. 
Figs.~\ref{ablation:module_traj}(b) and \ref{ablation:module_traj}(c) show a qualitative evaluation, demonstrating the usefulness of foreground information for camera localization.

\subsubsection{Human Scale Regularization}

\begin{figure}[!t]
\footnotesize
\centering
\begin{tabular}{c}
		\includegraphics[height=0.42\linewidth]{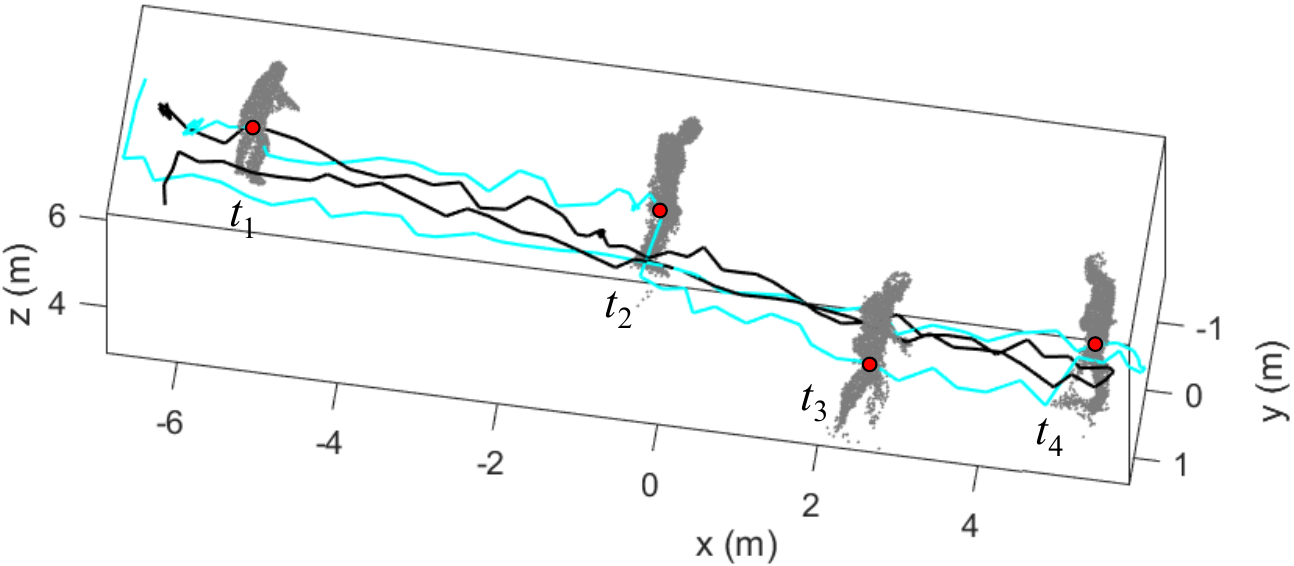} \\[-0.25em]
  (a)  \\
		\includegraphics[height=0.38\linewidth]{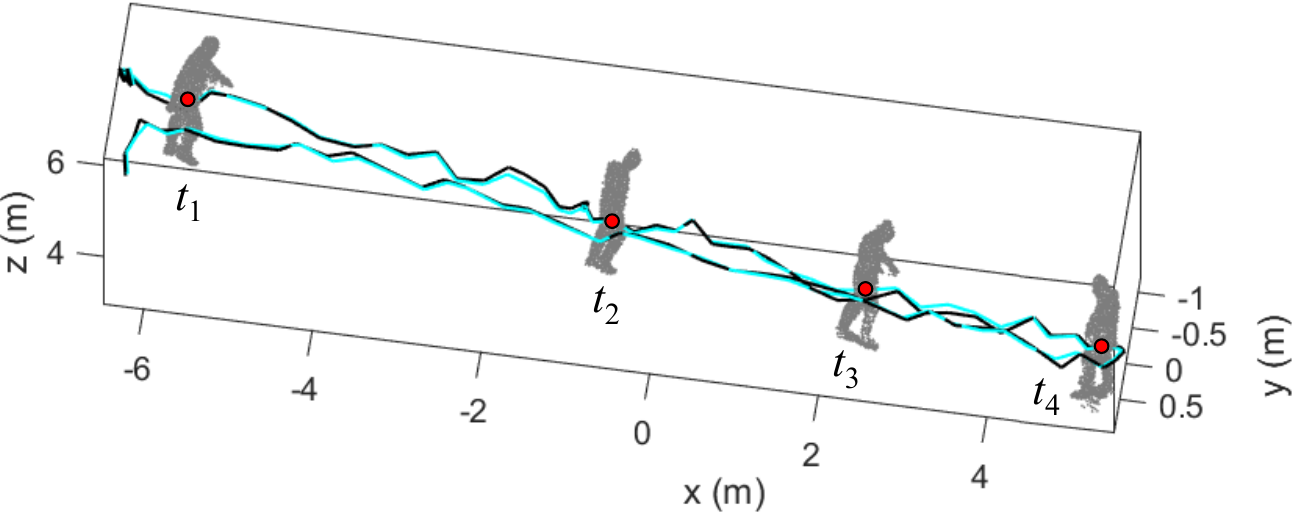}\\[-0.25em]
        
		  (b)  
	\end{tabular}\\
       	\caption{Ablation study of human scale regularization on Sequence \texttt{bike} of \textsf{NeuMan} dataset~\cite{neuman}. (a) Without regularization (b) With regularization. The notations are the same as those in Fig.~\ref{fig:human_traj}.} 
       	\label{ablation_scale}
\end{figure}

Recall that we apply a human scale regularization loss to human reconstruction in Section~\ref{subsec:mapping_human}.
This loss is designed to address the issue of low-quality depth images, which is particularly serious on \textsf{NeuMan} dataset. 
For validation, we compare our complete method to the version without this loss on \textsf{NeuMan} dataset. 
As shown in Table~\ref{table:human_traj}, this loss can effectively reduce the error of human trajectory on all the sequences.
A qualitative result in Fig.~\ref{ablation_scale} illustrates that this regularization can avoid inappropriate human sizes and floating Gaussians, and also mitigate the discontinuity of the estimated human trajectory.

\subsubsection{Optimization Between Neighboring Local Maps}

\begin{table}[!t]
	\renewcommand\arraystretch{1.3} 
	\renewcommand{\tabcolsep}{13pt} 
	\centering
 \caption{Ablation study of human scale regularization on \textsf{NeuMan} dataset~\cite{neuman}. We report the absolute trajectory error of human trajectory.}
		\begin{tabular}{c|cc|cc}
			\toprule
\multicolumn{1}{c}{Sequence}  &          
\multicolumn{2}{c}{Without Scale} & 
\multicolumn{2}{c}{With Scale}  
\\
\midrule
\textit{}  & RMSE & SD  &  RMSE & SD
\\
\texttt{bike}  & 21.1 & 9.9 &\textbf{6.0}  &\textbf{4.5}  \\
\texttt{citron}  & 19.0 & 11.2 & \textbf{14.2} &\textbf{8.7} \\
\texttt{jogging}  &  76.0  & 56.9 & \textbf{8.2} &\textbf{4.9}    \\
\texttt{parkinglot} & 18.6 & 7.3  & \textbf{15.2} &\textbf{6.7}  \\
\texttt{seattle} & 23.5 &  13.3 & \textbf{9.4} &\textbf{5.5}   \\
\midrule
Average & 36.3 &  19.7  & \textbf{10.6} &\textbf{6.0}   \\
\bottomrule 
			
\end{tabular}
\label{table:human_traj}
\end{table}

\begin{figure}[!t]
\footnotesize
\centering
	\renewcommand{\tabcolsep}{0.25pt} 
\begin{tabular}{ll}
     \includegraphics[height=0.375\linewidth]{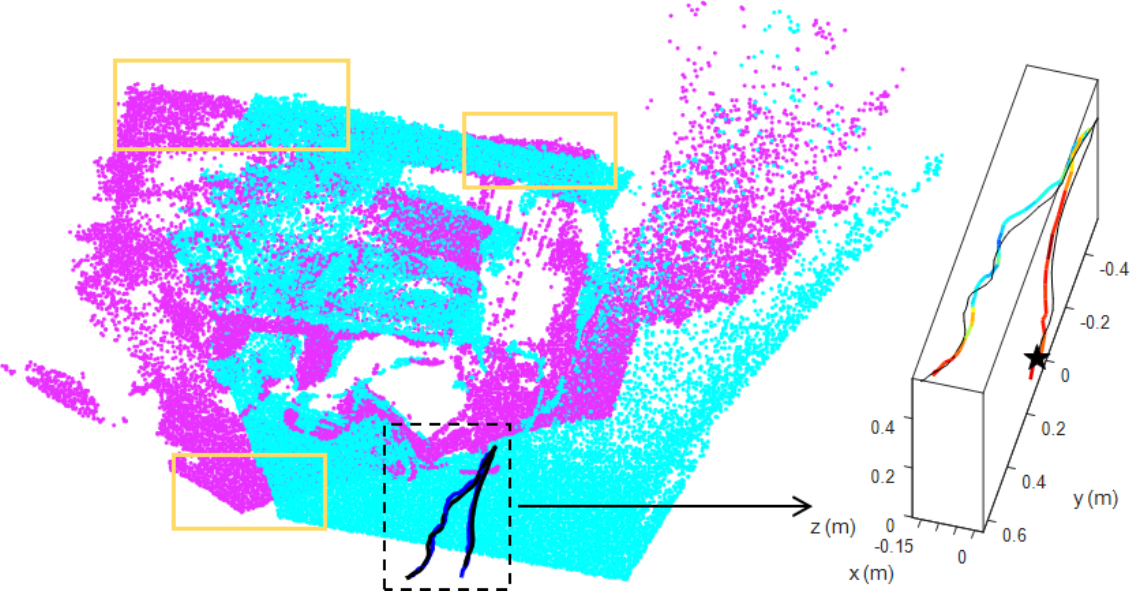} & \includegraphics[height=0.375\linewidth]{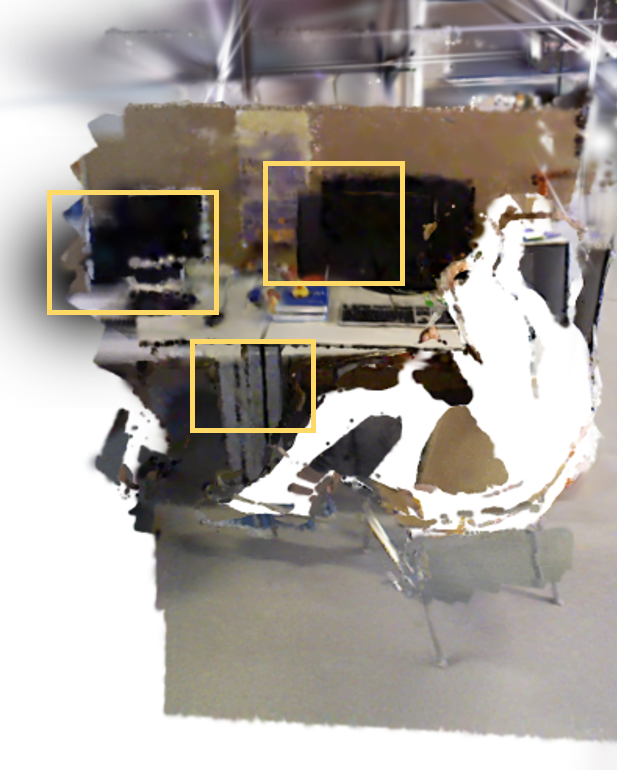}
     \\
     \multicolumn{2}{c}{(a) \textsf{Gaussian-SLAM}~\cite{gaussianslam}}   
     \\[0.5em]
     \includegraphics[height=0.375\linewidth]{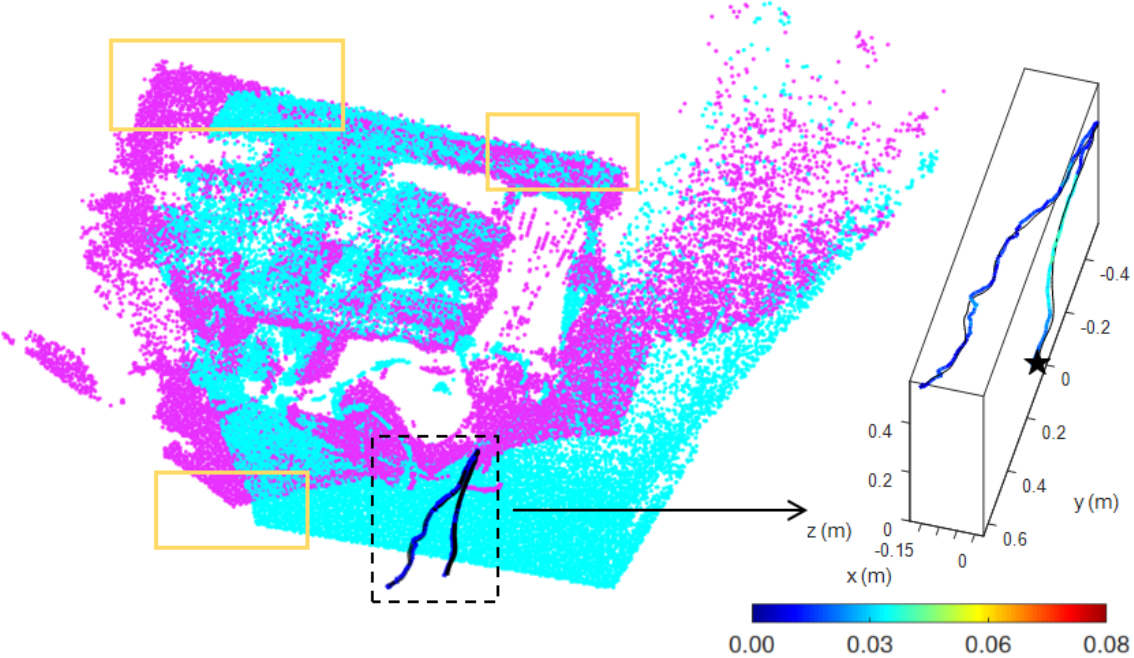} & 
     \includegraphics[height=0.375\linewidth]{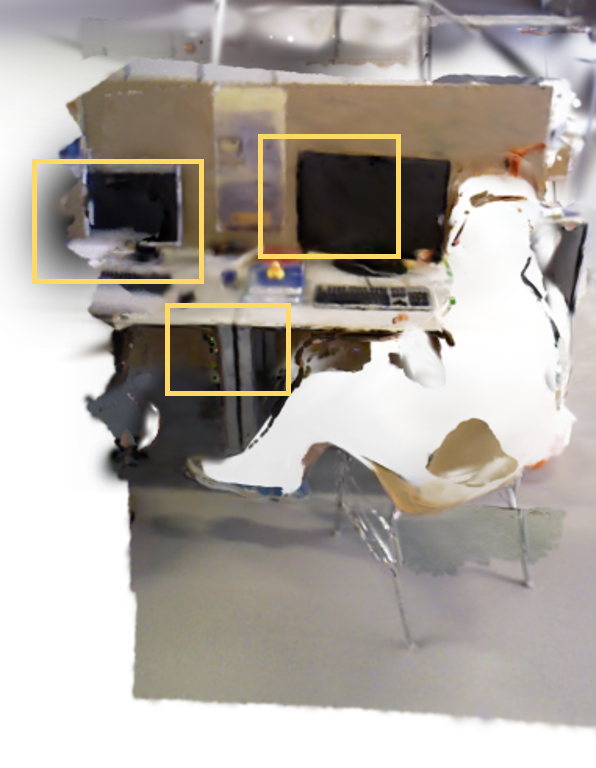}
     \\
     \multicolumn{2}{c}{(b) \textsf{PG-SLAM} (our)}  
\end{tabular}
       	\caption{Ablation study of optimization between neighboring local maps on sequence~\texttt{sitting\_halfsphere} of \textsf{TUM} dataset~\cite{tumbenchmark}. (a) Before optimization. (b) After optimization. Left: We report 
        representative
        local maps shown in magenta and cyan (for visualization, we show the centers of Gaussians). Middle: The colored and black lines denote the estimated and ground truth trajectories, respectively. Color bar indicates the magnitude of the absolute trajectory error. Pentagram represents the starting point of camera trajectory. Right: We present the rendered background from a novel view.}
       	\label{fig:abl_study_local_map}
\end{figure}

Recall that we introduce a strategy to optimize local maps in Section~\ref{subsec:opt_local_maps}.
We evaluate the effect of optimization on all the datasets, as shown in Table~\ref{table:submap}. On sequences of \textsf{TUM} dataset, the camera typically moves within the same scene throughout. Accordingly, the overlapping regions between neighboring local maps are relatively large.
In this case, the accuracy of both 3D map and camera trajectory is significantly improved based on our optimization strategy.
A representative result is shown in Fig.~\ref{fig:abl_study_local_map}.
By contrast, on \textsf{Bonn} and \textsf{NeuMan} datasets, the camera gradually explores new areas, leading to relatively small overlapping regions between adjacent local maps. Consequently, the effectiveness of the proposed optimization algorithm decreases.

\setlength{\tabcolsep}{3.5pt}
\begin{table}[!t]
\renewcommand\arraystretch{1.3}
\renewcommand{\tabcolsep}{13.5pt} 
\centering
\caption{Ablation study of optimization between neighboring local maps on three datasets. We report the absolute trajectory error of camera trajectory.}
\begin{tabular}{c|cc|cc}
			\toprule
			\multicolumn{1}{c}{Dataset}  &  \multicolumn{2}{c}{Without  Correction}& \multicolumn{2}{c}{With Correction}  \\
   
		\midrule
                \textit{}  & RMSE & SD  & RMSE & SD \\
			\textsf{TUM}~\cite{tumbenchmark}  & 6.2 &2.3 & \textbf{4.5}   & \textbf{1.8}  \\
                \textsf{Bonn}~\cite{ReFusion}  & 6.6 &2.4 & \textbf{6.5}   & \textbf{2.2}  \\
                \textsf{NeuMan}~\cite{neuman}  & 1.5 & 0.7 & \textbf{1.4}   & \textbf{0.6} \\
			\bottomrule 
			
		\end{tabular}
		\label{table:submap}
		
\end{table}
\section{Conclusions}

In this paper, we propose a photo-realistic and geometry-aware RGB-D SLAM method in dynamic environments.
To the best of our knowledge, our method is the first Gaussian splatting-based approach that can 
not only localize the camera and reconstruct the static background, but also map the dynamic humans and items. 
For foreground mapping, we estimate the deformations and/or motions of dynamic objects by considering the shape priors of humans and exploiting both geometric and appearance constraints with respect to Gaussians.
To map the background, we design an effective optimization strategy between neighboring local maps.
To localize the camera,
our method simultaneously uses the geometric and appearance constraints  by associating 3D Gaussians with 2D optical flows and pixel patches.
We leverage information of both static background and dynamic foreground to compensate for noise, effectively improving the localization accuracy.
Experiments on various real-world datasets demonstrate that our method outperforms state-of-the-art approaches.

\bibliographystyle{IEEEtran}
\bibliography{papers}

\end{document}